\documentclass[letterpaper,11pt]{article}

\usepackage{times}
\usepackage{diagbox}
\usepackage{makecell}
\usepackage{booktabs}
\usepackage{tikz}
\usepackage{amsmath}
\usepackage{amsfonts}
\usepackage{amssymb}
\usepackage{amsthm}
\usepackage{url,ifthen}
\usepackage{enumitem}
\usepackage{srcltx}
\usepackage{multirow}
\usepackage{boxedminipage}
\usepackage[margin=1in]{geometry}
\usepackage{nicefrac}
\usepackage{xspace}
\usepackage{graphicx}
\usepackage{color}
\usepackage{subfigure}
\usepackage{colortbl}
\usepackage{setspace}
\usepackage{natbib}
\usepackage{dsfont}
\usepackage{authblk}
\usepackage{soul}
\usepackage{longtable}
\setcitestyle{square, authoryear}
\definecolor{DarkGreen}{rgb}{0.1,0.5,0.1}
\definecolor{DarkRed}{rgb}{0.5,0.1,0.1}
\definecolor{DarkBlue}{rgb}{0.1,0.1,0.5}
\definecolor{Gray}{rgb}{0.2,0.2,0.2}

\usepackage{listings}
\lstdefinestyle{mystyle}{
    commentstyle=\color{DarkBlue},
    keywordstyle=\color{DarkRed},
    numberstyle=\tiny\color{Gray},
    stringstyle=\color{DarkGreen},
    basicstyle=\footnotesize,
    breakatwhitespace=false,         
    breaklines=true,                 
    captionpos=b,                    
    keepspaces=true,                 
    numbers=left,                    
    numbersep=5pt,                  
    showspaces=false,                
    showstringspaces=false,
    showtabs=false,                  
    tabsize=2
}
\usepackage[small]{caption}
\usepackage[pdftex]{hyperref}
\hypersetup{
    unicode=false,          
    pdftoolbar=true,        
    pdfmenubar=true,        
    pdffitwindow=false,      
    pdfnewwindow=true,      
    colorlinks=true,       
    linkcolor=DarkBlue,          
    citecolor=DarkGreen,        
    filecolor=DarkRed,      
    urlcolor=DarkBlue,          
    pdftitle={},
    pdfauthor={},
}

\usepackage[utf8]{inputenc}
\usepackage[T1]{fontenc}
\usepackage{microtype}
\usepackage{csquotes}
\usepackage{mathpazo}
\RequirePackage{helvet}
\RequirePackage{beramono}
\RequirePackage{textcomp}

\usepackage{macros}

\usepackage{thmtools,thm-restate}
\newtheorem{theorem}{Theorem}[section]

\newtheorem{lemma}[theorem]{Lemma}
\newtheorem{proposition}[theorem]{Proposition}
\theoremstyle{definition}
\newtheorem{definition}{Definition}[section]

\newtheorem*{remark}{Remark}

\newtheorem*{property*}{Property}

\usepackage{apptools}
\AtAppendix{\counterwithin{proposition}{section}}
\AtAppendix{\counterwithin{corollary}{section}}
\AtAppendix{\counterwithin{lemma}{section}}

\title{NICO$^{++}$: Towards Better Benchmarking for Domain Generalization}
\author{Xingxuan Zhang$^{\dag}$, Yue He$^{\dag}$, Renzhe Xu, Han Yu, Zheyan Shen, Peng Cui* \\
Department of Computer Science, Tsinghua University \\
~\\

{\small {xingxuanzhang@hotmail.com, heyue18@mails.tsinghua.edu.cn, xrz199721@gmail.com,}} \\
{\small {yuh21@mails.tsinghua.edu.cn, shenzy17@mails.tsinghua.edu.cn, cuip@tsinghua.edu.cn}}}
\date{}

\newcommand{\calA}{\mathcal{A}}
\newcommand{\calH}{\mathcal{H}}
\newcommand{\calD}{\mathcal{D}}
\newcommand{\calX}{\mathcal{X}}
\newcommand{\calY}{\mathcal{Y}}
\newcommand{\calM}{\mathcal{M}}
\newcommand{\calZ}{\mathcal{Z}}

\newcommand{\calG}{\mathcal{G}}
\newcommand{\calL}{\mathcal{L}}
\newcommand{\source}{{\text{tr}}}
\newcommand{\target}{{\text{te}}}
\newcommand{\bbR}{\mathbb{R}}
\newcommand{\bbE}{\mathbb{E}}
\newcommand{\bbI}{\mathbb{I}}
\newcommand{\dd}{\mathrm{disc}}
\newcommand{\concept}{\mathrm{cpt}}
\newcommand{\covariate}{\mathrm{cov}}
\newcommand{\rad}{\mathfrak{R}}
\newcommand{\kl}{D_{\mathrm{KL}}}

\makeatletter
\def\maketag@@@#1{\hbox{\m@th\normalfont\normalsize#1}}
\makeatother

\begin{document}
\maketitle
\begin{abstract}
\renewcommand{\thefootnote}{\fnsymbol{footnote}}
\footnotetext[2]{Equal contribution}
\footnotetext[1]{Corresponding Author}

Despite the remarkable performance that modern deep neural networks have achieved on independent and identically
distributed (I.I.D.) data, they can crash under distribution shifts.
Most current evaluation methods for domain generalization (DG) adopt the leave-one-out strategy as a compromise on the limited number of domains. We propose a large-scale benchmark with extensive labeled domains named NICO$^{++}$\footnote[3]{\scriptsize The dataset can be found at \url{https://www.dropbox.com/sh/u2bq2xo8sbax4pr/AADbhZJAy0AAbap76cg_XkAfa?dl=0}.

\setlength{\parindent}{2em}The Github repository for the paper is at \url{https://github.com/xxgege/NICO-plus}.} along with more rational evaluation methods for comprehensively evaluating DG algorithms. To evaluate DG datasets, we propose two metrics to quantify covariate shift and concept shift, respectively. Two novel generalization bounds from the perspective of data construction are proposed to prove that limited concept shift and significant covariate shift favor the evaluation capability for generalization. Through extensive experiments, NICO$^{++}$ shows its superior evaluation capability compared with current DG datasets and its contribution in alleviating unfairness caused by the leak of oracle knowledge in model selection.

\end{abstract}

\section{Introduction}

Machine learning has illustrated its excellent capability in a wide range of areas \citep{kipf2016semi,simonyan2014very,young2018recent}. Most current algorithms minimize the empirical risk in training data relying on the assumption that training and test data are independent and identically distributed (I.I.D.). However, this ideal hypothesis is hardly satisfied in real applications, especially those high-stake applications such as healthcare \citep{castro2020causality,miotto2018deep}, autonomous driving \citep{alcorn2019strike,dai2018dark,levinson2011towards} and security systems \citep{berman2019survey}, owing to the limitation of data collection and intricacy of the scenarios.
The distribution shift between training and test data may lead to the unreliable performance of most current approaches in practice. 
Hence, instead of generalization within the training distribution, the ability to generalize under distribution shift, namely domain generalization (DG) \citep{wang2021generalizing,zhou2021domain1}, is of more critical significance in realistic scenarios.

In the field of computer vision, benchmarks that provide the common ground for competing approaches often play a role of catalyzer promoting the advance of research \citep{deng2009imagenet}. An advanced DG benchmark should provide sufficient diversity in distributions for both training and evaluating DG algorithms \citep{xu2020neural,volpi2018generalizing} while ensuring essential common knowledge of categories for inductive inference across domains \citep{huang2020self,zhao2019learning,ilse2020diva}. The first property drives generalization challenging, and the second ensures the solvability \citep{ye2021ood}. This requires adequate distinct domains and instructive features for each category shared among all domains. 

Current DG benchmarks, however, either lack sufficient domains (e.g., 4 domains in PACS \citep{li2017deeper}, VLCS \citep{fang2013unbiased} and Office-Home \citep{venkateswara2017deep} and 6 in DomainNet \citep{peng2019moment}) or too simple or limited to simulating significant distribution shifts in real scenarios \citep{ganin2015unsupervised,arjovsky2019invariant,hendrycks2019benchmarking}. To enrich the diversity and perplexing distribution shifts in training data as much as possible, most of the current evaluation methods for DG adopt the leave-one-out strategy, where one domain is considered as test domain and the others for training. This is not an ideal evaluation for generalization but a compromise due to the limited number of domains in current datasets, which impairs the evaluation capability since the model is tested only on one specific distribution instead of multiple unseen distributions every time after training. 

To benchmark DG methods comprehensively and simulate real scenarios where a trained model may encounter any possible test data while providing sufficient diversity in the training data, we construct a large-scale DG dataset named NICO$^{++}$ with extensive domains and two protocols supported by aligned and flexible domains across categories, respectively, for better evaluation. Our dataset consists of 80 categories, 10 aligned common domains for all categories, 10 unique domains specifically for each category, and more than 200,000 images. Abundant diversity in both domain and category supports flexible assignments for training and test, controllable degree of distribution shifts, and extensive evaluation on multiple target domains. Images collected from real-world photos and consistency within category concepts provide sufficient common knowledge for recognition across domains on NICO$^{++}$.   

To evaluate DG datasets in depth, we investigate distribution shift on images (covariate shift) and common knowledge for category discrimination across domains (concept agreement) within them. Formally, we present quantification for covariate shift and the opposite of concept agreement, namely concept shift, via two novel metrics. We propose two novel generalization bounds and analyze them from the perspective of data construction instead of models. Through these bounds, we prove that limited concept shift and significant covariate shift favor the evaluation capability for generalization.   


Moreover, a critical yet common problem in DG is the model selection and the potential unfairness in the comparison caused by leveraging the knowledge of target data to choose hyperparameters that favors test performance \citep{gulrajani2020search,arpit2021ensemble}. This issue is exacerbated by the notable variance of test performance with various algorithm irrelevant hyperparameters on current DG datasets. Intuitively, strong and unstable concept shift such as confusing mapping relations from images to labels across domains embarrasses training convergence and enlarges the variance. 

We conduct extensive experiments on three levels. First, we evaluate NICO$^{++}$ and current DG datasets with the proposed metrics and show the superiority of NICO$^{++}$ in evaluation capability. Second, we conduct copious experiments on NICO$^{++}$ to benchmark current representative methods with the proposed protocols. Results show that the room for improvement of generalization methods on NICO$^{++}$ is spacious. Third, we show that NICO$^{++}$ helps alleviate the issue by squeezing the possible improvement space of oracle leaking and contributes as a fairer benchmark to the evaluation of DG methods, which meets the proposed metrics.

\section{Related Works}
In this section, we review the literature related to this paper, including benchmark datasets and domain generalization methods.

\paragraph{Benchmark Datasets.}
After the high-speed development benefited from the datasets, like PASCAL VOC \citep{everingham2015pascal}, ImageNet \citep{deng2009imagenet} and MSCOCO \citep{lin2014microsoft}, in IID scenarios, 
a range of image datasets have been raised for the research of domain generalization in visual recognition. 
The first branch modifies traditional image datasets with synthetic transformations, such as special data selection policies, perturbations or interventions, to simulate distribution shifts, typically including the ImageNet variants \citep{hendrycks2021many,hendrycks2019benchmarking,hendrycks2021natural}, MNIST variants \citep{arjovsky2019invariant,ghifary2015domain} and Waterbirds \citep{sagawa2019distributionally}. 
Another branch considers collecting data coming from different source domains, including PACS \citep{li2017deeper}, Office-Home \citep{venkateswara2017deep}, WILDS \citep{koh2021wilds}, DomainNet \citep{peng2019moment}, Terra Incognita \citep{beery2018recognition}, NICO \citep{he2021towards}, and VLCS \citep{fang2013unbiased}. 
In specific scenarios, Camelyon17 \citep{bandi2018detection} has tissue slides sampled and post-processed in different hospitals; FMoW \citep{christie2018functional} contains the satellites in distinct time and locations.
However, these datasets utilize a simple criterion to distinguish distributions, e.g. image style, not enough to cover the complexity in reality. In addition, the domains of most current DG datasets are limited, leading to inadequate diversity in training or test data.  
iWildCam \citep{beery2021iwildcam}, a large-scale dataset, takes pictures of wild animals with cameras at different locations and produces realistic distributional shifts. 
But it lacks the ability to control the strength of distribution shift to simulate diverse DG settings. 
The last version of NICO \citep{he2021towards} is insufficient to support some typical settings such as DA and DG since the domains are not aligned across categories.


\paragraph{Domain Generalization.}
There are several streams of literature studying the domain generalization problem in vision.
With extra information on test domains, domain adaptation methods  \citep{ben2006analysis,fang2020rethinking,ghafoorian2017transfer,sener2016learning,sugiyama2007covariate,sugiyama2007direct,tahmoresnezhad2017visual,xu2021gradual,zhang2016lsdt} show effectiveness in addressing the distribution shift problems.
By contrast, domain generalization aims to learn models that generalize well on unseen target domains while only data from several source domains are accessible.
According to \citep{shen2021towards}, DG methods can be divided into three branches, including representation learning \citep{blanchard2017domain,blanchard2011generalizing,gan2016learning,grubinger2015domain,jin2021style,muandet2013domain,nam2018batch,ghifary2016scatter,hu2020domain}, training strategies 	\citep{ding2017deep,wang2020dofe,segu2020batch,mancini2018best,zhang2021domain,liao2020deep,carlucci2019domain,ryu2019generalized,li2019episodic,huang2020self}, and data augmentation methods \citep{yue2019domain,tobin2017domain,peng2018sim,khirodkar2019domain,tremblay2018training,prakash2019structured,shankar2018generalizing,volpi2018generalizing,zhou2020deep}.
More comprehensive surveys on domain generalization methods can be found in \citep{wang2021generalizing,zhou2021domain}.



\section{NICO$^{++}$: Domain-Extensive Large Scale Domain Generalization Benchmark}
In this section, we introduce a novel large-scale domain generalization benchmark NICO$^{++}$, which contains extensive domains and categories. 
Similar to the original version of NICO \citep{he2021towards}, each image in NICO$^{++}$ consists of two kinds of labels, namely the category label and the domain label. The category labels correspond to the objective concept (\textit{e.g.}, cat and dog) while the domain labels represent other visual information (\textit{e.g.}, on grass, in water) in the images.
To boost the heterogeneity in the dataset to support the thorough evaluation of generalization ability in domain generalization scenarios,
we greatly enrich the types of categories and domains and collect a larger amount of images in NICO$^{++}$.  

\subsection{Constructions of the Category / Domain Labels}

We first select 80 categories and then build 10 common and 10 category-specific domains upon them. We provide detailed discussions on the selection of the categories and domains in Appendix.

\paragraph{Categories.}
Total 80 categories are provided with a hierarchical structure in NICO$^{++}$. Four broad categories \textit{Animal}, \textit{Plant}, \textit{Vehicle}, and \textit{Substance} lie on the top level. 
For each of \textit{Animal}, \textit{Plant}, and \textit{Vehicle}, there exist narrow categories derived from it (e.g., \textit{felida} and \textit{insect} belong to \textit{Animal}) in the middle level.
Finally, 80 concrete categories are assigned to their super-category respectively. 
The hierarchical structure ensures the diversity and balance\footnote{The ratio of the number of categories in \textit{Animal}, \textit{Plant}, \textit{Vehicle} and \textit{Substance} is $40:12:14:14$.} of categories in NICO$^{++}$, which is vital to simulate realistic domain generalization scenarios in wild environments. Detailed category structure is in Appendix.

\paragraph{Common domains.}
Towards the settings of domain generalization or domain adaption, we design $10$ common domains that are aligned across all categories.
Each of the selected common domains refers to a family of concrete contexts with similar semantics so that they are general and common enough to generate meaningful combinations with all categories.
For example, the common domain \textit{water} contains contexts of \textit{swimming}, \textit{in pool}, \textit{in river}, etc. Comparison between common domains in NICO$^{++}$ and domains in current DG datasets is in Appendix.

\paragraph{Unique domains.}
To increase the number of domains and support the flexible DG scenarios where the training domains are not aligned with respect to categories, we further attain unique domains specifically for each of the 80 categories. 
We select the unique domains according to the following conditions: 1) they are different from the common domains; 2) they can include various concepts, such as attributes (e.g. action, color), background, camera shooting angle, and accompanying objects, etc.; 3) different types of them hold a balanced proportion for diversity.

\subsection{Data Collection and Statistics}
\begin{figure*}[t]
    \centering
    \includegraphics[width=\linewidth]{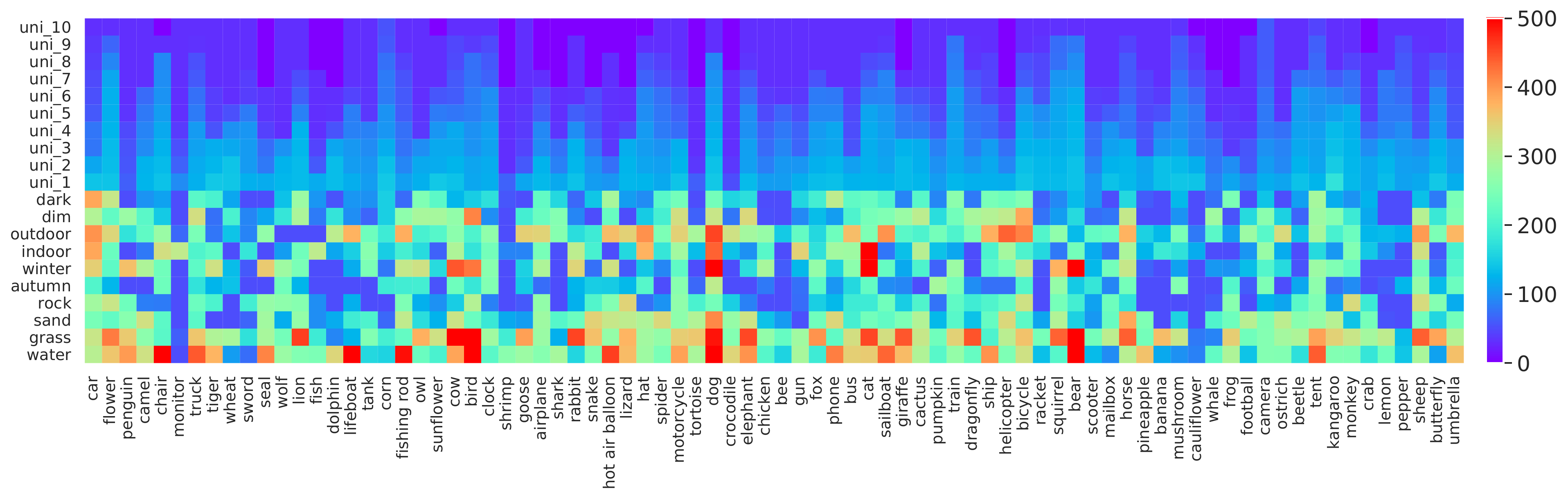}
    \caption{Statistical overview of NICO$^{++}$. The figure shows the number of instances in each domain and each category. The horizontal axis is for categories and the vertical axis for domains. The color of each bin corresponds to the number of instances in each \textit{(category, domain)} pair. The 10 domains at the bottom are common domains and identical for all categories, while the 10 at the top are unique domains that vary across categories and are represented with \{uni\_1, uni\_1, ..., uni\_10\}.}
    \figurelabel{fig:statistic}
\end{figure*}

NICO$^{++}$ has 10 common domains, covering nature, season, humanity and illumination, for total 80 categories, and 10 unique domains for each category. The capacity of most common domains and unique domains is at least 200 and 50, respectively. 
The images from most domains are collected by searching a combination of a category name and a phrase extend from the domain name (e.g. ``dog sitting on grass" for the category \textit{dog} and the domain \textit{grass}). Over 32,000 combinations are adopted for searching images. The downloaded data contain a large portion of outliers that require artificial annotations. Each image is assigned to two annotators and passes the selection when agreed by both annotators. After the annotation process, 232.4k images are selected from over 1.0 million images downloaded from the search engines.

The scale of NICO$^{++}$ is enormous enough to support the training of deep convolutional networks (\textit{e.g.}, ResNet-50) from scratch in types of domain generalization scenarios. A statistical overview of the dataset is shown \figureref{fig:statistic}.

\section{Covariate Shift and Concept Shift}
\sectionlabel{sec:evaluation}

Consider a dataset with data points sampled from a joint distribution $P(X,Y) = P(Y|X)P(X)$. Distribution shift within the dataset can be caused by the shift on $P(X)$ (\textit{i.e.}, covariate shift) and shift on $P(Y|X)$ (\textit{i.e.}, concept shift) \citep{shen2021towards}. We give quantification for these two shifts in any labeled dataset and analyze the preference of them from a perspective of the DG benckmark via presenting two generalization bounds for multi-class classification. Then we evaluate NICO$^{++}$ and current DG datasets empirically with the proposed metrics and show the superiority of NICO$^{++}$.

\paragraph{Notations} We use $\calX$ and $\calY$ to denote the space of input $X$ and outcome $Y$, respectively. We use $\Delta_{\calY}$ to denote a distribution on $\calY$. A domain $d$ corresponds to a distribution $\calD_d$ on $\calX$ and a labeling function\footnote{We use $\Delta_\calY$ here to denote that the labeling function may not be deterministic. This formulation also includes deterministic labeling function cases.} $f_d: \calX \rightarrow \Delta_\calY$. The training and test domains are specified by $(\calD_{\source}, f_{\source})$ and $(\calD_\target, f_\target)$, respectively. We use $p_\source(x)$ and $p_\target(x)$ to denote the probability density function on training and test domains.
Let $\ell: \Delta_\calY \times \Delta_\calY \rightarrow \bbR_{+}$ define a loss function over $\Delta_\calY$ and $\calH$ define a function class mapping $\calX$ to $\Delta_{\calY}$. For any hypotheses $h_1, h_2 \in \calH$, the expected loss $\calL_{\calD}(h_1, h_2)$ for distribution $\calD$ is given as $\calL_{\calD}(h_1, h_2) = \bbE_{x \sim \calD}\left[\ell(h_1(x), h_2(x))\right]$. To simplify the notations, we use $\calL_\source$ and $\calL_\target$ to denote the expected loss $\calL_{\calD_\source}$ and $\calL_{\calD_\target}$ in training and test domain, respectively. In addition, we use $\varepsilon_{\source}(h) = \calL_{\source}\left(h, f_{\source}\right)$ and $\varepsilon_{\target}(h) = \calL_{\target}\left(h, f_{\target}\right)$ to denote the loss of a function $h \in \calH$ \textit{w.r.t.} to the true labeling function $f_\source$ and $f_\target$, respectively.

\subsection{Metrics for Covariate shift and Concept shift}
\sectionlabel{sec:metrics}
The distribution shift between the training domain $(\calD_\source, f_\source)$ and test domain $(\calD_\target, f_\target)$ can be decomposed into covariate shift (\textit{i.e.}, shift between $\calD_\source$ and $\calD_\target$) and concept shift (\textit{i.e.}, shift between $f_\source$ and $f_\target$). We propose the following metrics to measure the covariate shift and concept shift.
\begin{definition} [Metrics for covariate shift and concept shift]
    Let $\calH$ be a set of functions mapping $\calX$ to $\Delta_{\calY}$ and let $\ell: \Delta_{\calY} \times \Delta_{\calY} \rightarrow \bbR_+$ define a loss function over $\Delta_\calY$. For the two domains $(\calD_\source, f_\source)$ and $(\calD_\target, f_\target)$, then
    \begin{itemize}[leftmargin=*,noitemsep,topsep=0pt,parsep=0pt,partopsep=0pt]
        \item the covariate shift is measured as the discrepancy distance \citep{mansour2009domain} (provided in \definitionref{defn:dd}) between $\calD_\source$ and $\calD_\target$ under $\calH$ and $\ell$, \textit{i.e.},
        \begin{equation}
            \calM_{\covariate}\left(\calD_\source, \calD_\target; \calH, \ell\right) \triangleq \dd\left(\calD_\source, \calD_\target; \calH, \ell\right),
        \end{equation}
        \item the concept shift is measured as the maximum / minimum loss when using $f_\source$ on the test domain or using $f_\target$ on the training domain, \textit{i.e.},
        \begin{equation}
            \left\{
            \begin{aligned}
                \calM_{\concept}^{\min}\left(\calD_\source, \calD_\target, f_\source, f_\target; \ell\right) & \triangleq \min\left\{\calL_{\source}(f_{\source}, f_{\target}), \calL_{\target}(f_{\source}, f_{\target})\right\}, \\ \calM_{\concept}^{\max}\left(\calD_\source, \calD_\target, f_\source, f_\target; \ell\right) & \triangleq \max\left\{\calL_{\source}(f_{\source}, f_{\target}), \calL_{\target}(f_{\source}, f_{\target})\right\}.
            \end{aligned}
            \right.
        \equationlabel{eqn:concept}
        \end{equation}
    \end{itemize}
\end{definition}

\begin{remark}
    We introduce two metrics for concept shift terms in \equationref{eqn:concept} because they both provide meaningful characterizations of the concept shift. In addition, both $\calM_{\concept}^{\min}$ and $\calM_{\concept}^{\max}$ have close connections with DG performance as shown in \theoremref{thrm:bound-population} and \theoremref{thrm:bound-population-minus} in \sectionref{sect:inequalities}. The covariate shift is widely discussed in recent literature \citep{duchi2020distributionally,ruan2021optimal,shen2021towards} yet none of them give the quantification with function discrepancy, which favors the analysis of DG performance and shows remarkable properties when $\calH$ is large (such as the function space supported by current deep models). 
    The concept shift can be considered as the discrepancy between the labeling rule $f_{\source}$ on the training data and the labeling rule $f_{\target}$ on the test data. Intuitively, consider that a circle in the training data is labeled as class \textit{A} in training domains and class \textit{B} in test domains, models can hardly learn the labeling function on the test data (mapping the circle to class \textit{B}) without knowledge about test domains.
\end{remark}

The discrepancy distance mentioned above is defined as follows.
\begin{definition} [Discrepancy Distance~\citep{mansour2009domain}] \definitionlabel{defn:dd}
    Let $\calH$ be a set of functions mapping $\calX$ to $\Delta_{\calY}$ and let $\ell: \Delta_{\calY} \times \Delta_{\calY} \rightarrow \bbR_+$ define a loss function over $\Delta_\calY$. The discrepancy distance $\dd\left(\calD_1, \calD_2; \calH, \ell\right)$ between two distributions $\calD_1$ and $\calD_2$ over $\calX$ is defined by
    \begin{equation}
        \dd\left(\calD_1, \calD_2; \calH, \ell\right) \triangleq \sup_{h_1, h_2 \in \calH}\left|\calL_{\calD_1}(h_1, h_2) - \calL_{\calD_2}(h_1, h_2)\right|.
    \end{equation}
\end{definition}

We give formal analysis of metrics for covariate shift ($\calM_{\covariate}$) and concept shift ($\calM_{\concept}^{\min}$/$\calM_{\concept}^{\max}$) below and the graphical explanation is shown in \figureref{fig:metrics}.

\begin{figure}[t]
    \centering
    \includegraphics[width=0.8\linewidth]{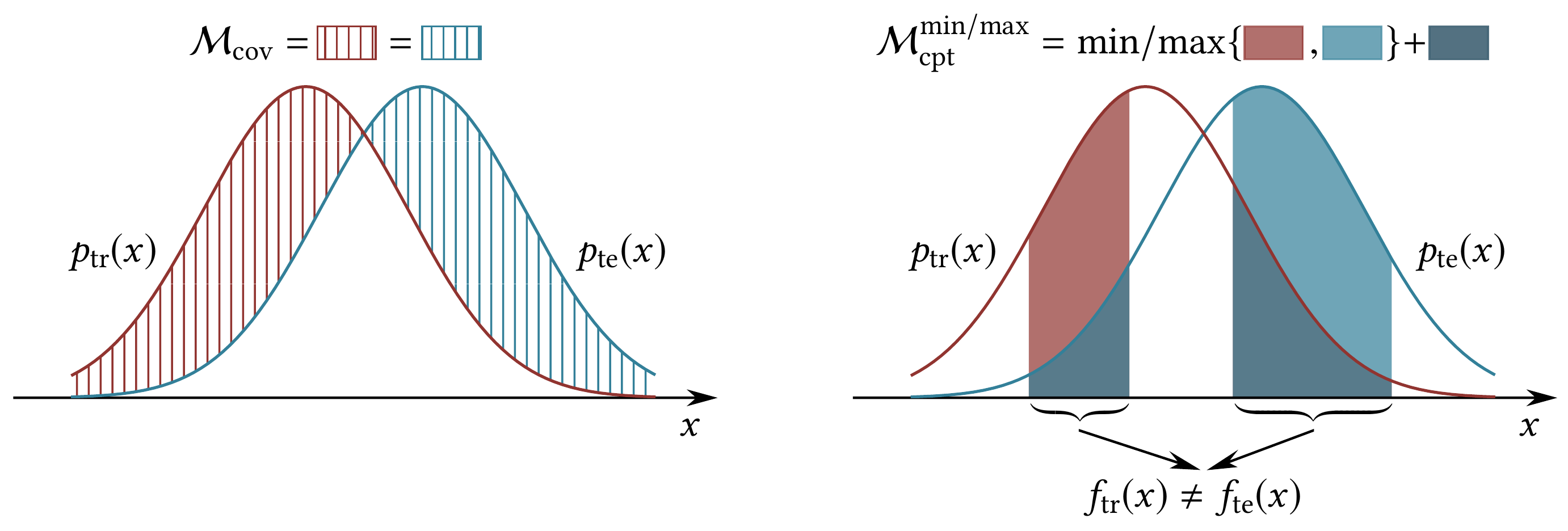}
    \caption{Graphical explanations of our proposed metric $\calM_{\covariate}$ and $\calM_{\concept}^{\min}$/$\calM_{\concept}^{\max}$ when $\calH$ is the set of all functions mapping $\calX$ to $\Delta_\calY$ and $\ell$ is the 0-1 loss.}
    \figurelabel{fig:metrics}
\end{figure}

\paragraph{The covariate shift term $\calM_{\covariate}$.}
When the capacity of function class $\calH$ is large enough and $\ell$ is bounded, $\calM_{\covariate}$ is in terms of the $\ell_1$ distance between two distributions, given by the following proposition.
\begin{proposition} \propositionlabel{prop:dd}
    Let $\calH$ be the set of all functions mapping $\calX$ to $\Delta_\calY$ and the range of the loss function is $[0, M]$, then for any two distributions $\calD_{\source}$ and $\calD_\target$ on $\calX$ with probability density function $p_\source$ and $p_\target$ respectively,
    \begin{equation}
        \calM_{\covariate}\left(\calD_\source, \calD_\target; \calH, \ell\right) = \frac{M}{2}\ell_1\left(\calD_\source, \calD_\target\right) = \frac{M}{2}\int_{\calX}\left|p_\source(x) - p_\target(x)\right|\mathrm{d}x.
    \end{equation}
\end{proposition}

It is clear that the covariate shift metric $\calM_{\covariate}$ is determined by the accumulated bias between the distribution $\calD_{\source}$ and $\calD_{\target}$ defined on $\calX$ and without contribution from $\calY$, which meets the definition of covariate shift.

\paragraph{The concept shift term $\calM_{\concept}^{\min}$ and $\calM_{\concept}^{\max}$.}
When $\ell$ is set as the 0-1 loss, \textit{i.e.}, the loss $\ell(f_\source(x), f_{\target}(x))$ is $0$ if and only if $f_\source(x) = f_\target(x)$, $\calM_{\concept}^{\min}$ and $\calM_{\concept}^{\max}$ can be written as follows.
\begin{equation}
    \begin{aligned}
        \calM_{\concept}^{\min} & =  \min\left\{\int_{\calX}\bbI[f_{\source}(x) \ne f_{\target}(x)]p_{\source}(x)\mathrm{d}x, \int_{\calX}\bbI[f_{\source}(x) \ne f_{\target}(x)]p_{\target}(x)\mathrm{d}x\right\} \\
        \calM_{\concept}^{\max} & =  \max\left\{\int_{\calX}\bbI[f_{\source}(x) \ne f_{\target}(x)]p_{\source}(x)\mathrm{d}x, \int_{\calX}\bbI[f_{\source}(x) \ne f_{\target}(x)]p_{\target}(x)\mathrm{d}x\right\} \\
    \end{aligned}
\end{equation}
Here $\bbI[f_\source(x) \ne f_\target(x)]$ is an indicator function on whether $f_\source(x) \ne f_\target(x)$. Intuitively, the two terms in the $\min$/$\max$ functions represent the probabilities of inconsistent labeling function in training and test domains.
$\calM_{\concept}^{\min}$ and $\calM_{\concept}^{\max}$ further take the minimal and maximal value of the two probabilities, respectively. It is rational that the concept shift is actually the integral of $p_{\source}(x)$ (or $p_{\target}(x)$) over any points $x$ where its corresponding label on training data differs from that on test data.
In practice, we estimate $f_{\source}$ and $f_{\target}$ with models trained on source domains and target domains, respectively. More discussion and comparison of discrepancy distance and other metrics for distribution distance is in Appendix.

\subsection{Dataset Evaluation with the Metrics} \sectionlabel{sect:inequalities}
To use the covariate shift metric $\calM_{\covariate}$ and concept shift metrics $\calM_{\concept}^{\min}, \calM_{\concept}^{\max}$ for dataset evaluation, we show that larger covariate shift and smaller concept shift favors a discriminative domain generalization benchmark. 
Intuitively, the critical point of datasets for domain generalization lies in 1) significant covariate shift between domains that drives generalization challenging \citep{quinonero2008dataset} and 2) common knowledge about categories across domains on which models can rely on to conduct valid predictions on unseen domains \citep{zhao2019learning,ilse2020diva}. The common knowledge requires the alignment between labeling functions of source domains and target domains, \textit{i.e.}, a moderate concept shift. When there is a strong inconsistency between labeling rules on training and test data, the classification loss instructing biased connections between visual features and concepts is misleading for generalization to test data. Thus models can hardly learn strong predictors for test data without knowledge of test domain. 

To analyze the intuitions theoretically, we first propose an upper bound for the expected loss in the test domain for any hypothesis $h \in \calH$.

\begin{theorem}
\theoremlabel{thrm:bound-population}
    Suppose the loss function $\ell$ is symmetric and obeys the triangle inequality. Suppose $f_{\source}, f_{\target} \in \calH$. Then for any hypothesis $h \in \calH$, the following holds
    \begin{equation} \equationlabel{eq:DA-bound}
        \varepsilon_{\target}(h) \le \varepsilon_{\source}(h) +  \calM_{\covariate}\left(\calD_\source, \calD_\target; \calH, \ell\right) + \calM_{\concept}^{\min}\left(\calD_\source, \calD_\target, f_\source, f_\target; \ell\right).
    \end{equation}
\end{theorem}

\begin{remark}
    \theoremref{thrm:bound-population} is closely related to generalization bounds in domain adaptation (DA) literature \citep{ben2006analysis,zhang2019bridging,zhao2019learning,zhang2020unsupervised}. In detail, \citep{ben2006analysis} first studied the generalization bound from a source domain to a target domain in binary classification problems and \citep{zhang2019bridging,zhang2020unsupervised} further extended the results to multi-class classification problems. However, the bounds in their results depend on a specific term $\lambda^* \triangleq \min_{h \in \calH} \varepsilon_{\source}(h) + \varepsilon_{\target}(h)$, which is conservative and relatively loose and can not be measured as concept shift directly \citep{zhao2019learning}. As a result, \citep{zhao2019learning} developed a bound which explicitly takes concept shift (termed as conditional shift by them) into account. However, their results are only applied to binary classifications and $\ell_1$ loss function. By contrast, \theoremref{thrm:bound-population} can be applied to multi-class classifications problems and any loss functions that are symmetric and obeys the triangle inequality.
\end{remark}

\theoremref{thrm:bound-population} quantitatively gives an estimation about the biggest gap between the performance of a model on training and test data. If we consider $\calH$ as a set of deep models trained on training data with different learning strategies, the estimation indicates the upper bound of range in which their performance varies.
If we consider $h$ as a model that fits training data, the bound gives an estimation of how much the distribution shift of the dataset contributes to the performance drop between training and test data.

Furthermore, we propose a lower bound for the expected loss in the test domain for any hypothesis $h \in \calH$ to better understand how the proposed metrics $\calM_{\concept}$ and $\calM_{\covariate}$ affects discrimination ability of datasets. 

\begin{theorem} \theoremlabel{thrm:bound-population-minus}
    Suppose the loss function $\ell$ is symmetric and obeys the triangle inequality. Suppose $f_{\source}, f_{\target} \in \calH$. Then for any hypothesis $h \in \calH$, the following holds
    \begin{equation} \equationlabel{eq:DA-bound-minus}
        \begin{aligned}
            \varepsilon_{\target}(h) \ge \calM_{\concept}^{\max}\left(\calD_\source, \calD_\target, f_\source, f_\target; \ell\right) - \calM_{\covariate}\left(\calD_\source, \calD_\target; \calH, \ell\right) - \varepsilon_{\source}(h).
        \end{aligned}
    \end{equation}
\end{theorem}

As shown in \theoremref{thrm:bound-population-minus}, for any hypothesis $h \in \calH$, the term ($\calM_{\concept}-\calM_{\covariate}$) determines the lower bound of the test loss and further determines the upper bound of the test performance of $h$. The bound is critical to evaluate a dataset since the performance of any well-trained model on test data is upper bounded by the properties (concept shift and covariate shift) of the dataset, disregarding how the model is designed or learned. Specifically, consider the stop training condition of a any possible model $h$ is that the loss on the training data is smaller than $\gamma$, which is rational with most of current training strategies, the performance of the model on test data is upper bounded by $\gamma - \calM_{\concept}+ \calM_{\covariate}$, which is irrelevant to the choice of $h$ and the learning protocol. Intuitively, when the discrepancy between labeling functions between training and test data, the better the model fits training data, the worse it generalizes to test domains. Conversely, with more aligned labeling functions, the common knowledge between training and test data is richer and more instructive, so that the ceiling of generalization is higher.
Moreover, the covariate shift $\calM_{\covariate}$ contributes positively to the upper bound of the test performance, given that the concept shift $\calM_{\concept}$ can be considered as integral of probability density $p_{\source}(x)$ (or $p_{\target}(x)$) over points with unaligned labeling functions, where the covariate shift $\calM_{\covariate}$ helps to counteract the impact of labeling mismatch.

As a result, the drop given by \theoremref{thrm:bound-population-minus} is unsolvable for algorithms but modifiable by suppressing the concept shift or enhancing the covariate shift. To better evaluate generalization ability, an DG benchmark requires small concept shift and large covariate shift. The empirical versions of \theoremref{thrm:bound-population} and \theoremref{thrm:bound-population-minus} are provided in Appendix.

\subsection{Empirical Evaluation}
\begin{table*}[t]
    \centering
    \caption{Results of estimated covariate shift and concept shift of NICO$^{++}$ and current DG datasets. $\uparrow$ donates that the higher the metric is, the better and $\downarrow$ is the opposite. The best results of all datasets are highlighted with the bold font.}
    \resizebox{\linewidth}{!}{
    \begin{tabular}{cccccccc}
        \toprule
         & I.I.D. & PACS & DomainNet & VLCS & Office-Home & MNIST-M & NICO$^{++}$  \\
        \midrule
        $\calM_{\covariate} \uparrow$ & 0 & 0.325\scriptsize{($\pm$0.053)} &  0.302\scriptsize{($\pm$0.039)} & 
        0.256\scriptsize{($\pm$0.041)}
        &
        0.238\scriptsize{($\pm$0.049)}
        & 
        0.225\scriptsize{($\pm$0.034)}
        & \textbf{0.338}\scriptsize{($\pm$0.031)}\\
        $\calM_{\concept}^{\min} \downarrow$ & 0 & 0.434\scriptsize{($\pm$0.023)} & 0.247\scriptsize{($\pm$0.055)} & 
        0.303\scriptsize{($\pm$0.064)}
        & 
        0.353\scriptsize{($\pm$0.086)}
        & 
        0.243\scriptsize{($\pm$0.048)}
        & \textbf{0.152}\scriptsize{($\pm$0.034)} \\
        $\calM_{\concept}^{\max} \downarrow$ & 0 & 0.537\scriptsize{($\pm$0.054)} & 0.612\scriptsize{($\pm$0.057)} & 
        0.523\scriptsize{($\pm$0.044)}
        & 
        0.505\scriptsize{($\pm$0.084)}
        & 
        0.449\scriptsize{($\pm$0.030)}
        & \textbf{0.192}\scriptsize{($\pm$0.040)}\\
        \bottomrule
    \end{tabular}}
    \tablelabel{tab:disc}
\end{table*}

We compare NICO$^{++}$ with current DG datasets in both covariate shift $\calM_{\covariate}$ and concept shift $\calM_{\concept}^{\min}$, $\calM_{\concept}^{\max}$.

For the covariate shift term, we first train two models from scratch jointly by optimizing the following two objective function, namely
\begin{equation}
    \calL_\dd^{(1)} = \calL_{\calD_\source}(h_1, h_2) - \calL_{\calD_\target}(h_1, h_2), \quad \calL_\dd^{(2)} = \calL_{\calD_\source}(h_1, h_2) - \calL_{\calD_\target}(h_1, h_2).
\end{equation}
We take the bigger one of the absolute value of $\calL_\dd^{(1)}$ and $\calL_\dd^{(2)}$ as the final indicator for covariate shift $\calM_{\covariate}$.
We adopt raw ResNet50 \citep{he2016deep} as the model for NICO$^{++}$, PACS, DomainNet, VLCS, and Office-Home and shallower CNNs (the structure is shown in Appendix) for MNIST-M \citep{ganin2015unsupervised} as its image size is small. For a fair comparison, we randomly select 2 domains as the source and 2 domains as the target for all datasets.
Since there are only 5 categories in VLCS, we randomly select 5 categories from each domain for each run and report the average of 5 runs. 
Source and target domains from different datasets are set to approximately the same capacity of images.
The learning rate for all models is set to 0.1, batch size is 64, and the number of training epoch is 20.

For the concept shift, we estimate $f_{\source}$ and $f_{\target}$ with models that fit the source set and target set, respectively. Specifically, we learn two models on the source and target set of a given dataset, respectively, with the objective of category recognition and each of them on both source and target data. More details of implementation can be found in Appendix.

Results are shown in \tableref{tab:disc}. Concept shift on NICO$^{++}$ is significantly lower than other datasets, indicating more aligned labeling rules across domains and more instructive common knowledge of categories can be learned by models. The covariate shifts of NICO$^{++}$, PACS, and DomainNet are comparable, which demonstrates that the distribution shift on images caused by the background can be as strong as style shifts. It is worthy to notice that the term $\calM_{\concept} - \calM_{\covariate}$ in \theoremref{thrm:bound-population-minus} is larger than 0 on current DG datasets while lower than 0 on NICO$^{++}$, indicating that the drop caused by a shift of labeling function across domains is significant enough to damage the upper generalization bound while the common knowledge across domains in NICO$^{++}$ is sufficient for models to approach the oracle performance.

\section{Experiments}
Inspired by \citep{zhang2021deep}, we present two evaluation settings, namely \textit{classic domain generalization} and \textit{flexible domain generalization} and perform extensive experiments on both settings.
We design experimental settings to evaluate current DG methods on NICO$^{++}$ and illustrate how NICO$^{++}$ contributes to filling in the evaluation on generalization to multiple unseen domains. Due to space limitations, we only report major results, and more experiments and implementation details are provided in Appendix.

\subsection{Evaluation Metrics for Algorithms}
Despite the fact that the widely adopted evaluation methods in DG effectively shows the generalization ability of models to the unseen target domain, they fail to sufficiently simulate real scenarios in application. For example, the most popular evaluation method, namely leave-one-out evaluation \citep{li2017deeper,shen2021towards}, tests models on a single target domain for each training process, while in real applications, a trained model is required to be reliable under any possible scenarios with various data distributions.
The compromise on the limitation of domain numbers in current benchmarks, including PACS, VLCS, DomainNet, Office-Home, can be addressed by NICO$^{++}$ with sufficient aligned and unique domains. The superiority supports designing more realistic evaluation metrics to test models' generalization ability comprehensively.

We consider three simple metrics to evaluate DG algorithm, namely average accuracy, overall accuracy, and the standard deviation of accuracy across domains. The metrics are defined as follows.
\begin{equation}
    \begin{aligned}
        & \mathrm{Average}  = \frac{1}{K} \sum_{k=1}^{K} \mathrm{acc}_k, \quad \mathrm{Overall} = \frac{1}{\sum_{k=1}^K N_k}\sum_{k=1}^K N_k \mathrm{acc}_k, \\
        & \mathrm{Std} = \sqrt{\frac{1}{K - 1}\sum_{k=1}^K(\mathrm{acc}_k - \mathrm{Average})^2}.
    \end{aligned}
\end{equation}
Here $K$ is the number of domains in the test data, $N_k$ is the number of samples in the $k$-th domain, and $\mathrm{acc}_k$ is the prediction accuracy in the $k$-th domain. The metric $\mathrm{Average}$ is widely used in the literature of DG, where both training and test domains for different categories are aligned. The metric $\mathrm{Overall}$ is more reasonable when the domains can be various for different categories or the test data are a mixture of unknown domains, and thus the accuracy for each domain is incalculable. The metric $\mathrm{Std}$ indicates the standard deviation of the performance across different domains. Since learning models that are consistently reliable in any possible environment is the target of DG and many methods are designed to learn invariant representations \citep{ganin2016domain}, $\mathrm{Std}$ is rational and instructive.  Please note that $\mathrm{Std}$ is insignificant in the leave-one-out evaluation method where models tested on different target domains are trained on different combinations of source domains, while domains of NICO$^{++}$ are rich enough to evaluate models on various target domains with fixed source domains.

\subsection{Benchmark for Classic Domain Generalization}

The common domains in NICO$^{++}$ are consistent for all categories, which supports the experimental designs of DG with aligned domains. 
They can be further grouped into 3 clusters with respect to the kind of distribution shift (detailed discussions are in Appendix), namely location (e.g., indoor or outdoor), background (e.g., around water or on grass), and time (e.g., dim or dark, winter or autumn) shift. In this section we consider two levels of distribution shift, where domains across clusters are selected for test and domains within the same cluster for test, respectively. Six domains are selected for training and the others for test and the results of current representative methods with ResNet-50 as the backbone are shown in \tableref{tab:dg}. Models generally show better generalization when tested on a single cluster of common domains than the opposite, indicating that generalization to diverse unseen domains is more challenging. Current SOTA methods such as EoA, CORAL, and StableNet show their effectiveness, yet a significant gap between them and the oracle performance shows that the room for improvement is spacious. More splits and implementation details are in Appendix.

\begin{table*}[t]
    \centering
    \caption{Results of the DG setting on NICO$^{++}$. We report the accuracy on each target domain, overall accuracy, mean accuracy, and variance of accuracies across all target domains. We reimplement state-of-the-art unsupervised methods on DomainNet with ResNet-50 \citep{he2016deep} as the backbone network for all the methods unless otherwise specified. Oracle donates the ResNet-50 trained with data sampled from the target distribution (yet none of test images is seen in the training). Ova. and Avg. indicate the overall accuracy of all the test data and the arithmetic mean of the accuracy of 3 domains, respectively. Note that they are different because the capacities of different domains are not equal. The reported results are average over three repetitions of each run. The best results of all methods are highlighted with the bold font and the second best with underlined font.}
    \resizebox{1\textwidth}{!}{
    \begin{tabular}{c|cccc|ccc|cccc|ccc}
        \toprule
        \multirow{2}{*}{Method} & \multicolumn{7}{c|}{Training domains: G, Wa, R, A, I, Di} & \multicolumn{7}{c}{Training domains: S, G, Wa, R, I, O} \\
        \cmidrule{2-15}
        & S & Wi & O & Da & Ova. & Avg. & Std & A & Wi & Da & Di & Ova. & Avg. & Std \\
        \midrule
        Deepall &   80.95 & 79.96 & 73.30 & 76.27 & 77.50 & 77.62 & 3.05 & 81.47 & 79.53 & 78.13 & 77.19 & 79.20 & 79.08 & 1.61 \\ 
        SWAD \citep{cha2021swad} & \textbf{82.71} & \textbf{81.92} & 76.15 & 77.20 & \textbf{79.54} & \textbf{79.50} & 2.86 & \textbf{82.95} & 80.33 & 79.16 & 77.58 & 79.82 & 80.00 & 1.96 \\ 
        MMLD \citep{matsuura2020domain} & 76.45 & 80.11 & 76.25 & 76.91 & 77.40 & 77.43 & \textbf{1.57} & 80.25 & 78.27 & 78.56 & 76.23 & 78.15 & 78.33 & 1.43 \\ 
        RSC \citep{huang2020self} &   80.07 & 80.22 & \textbf{76.67} & 76.14 & 78.37 & 78.27 & 1.88 & 81.22 & 80.61 & 78.45 & 77.60 & 79.42 & 79.47 & 1.49 \\
        AdaClust \citep{thomas2021adaptive} & 79.57 & 78.53 & 71.75 & 74.91 & 76.06 & 76.19 & 3.09 & 80.40 & 78.63 & 76.53 & 75.82 & 77.96 & 77.85 & 1.80 \\
        SagNet \citep{nam2021reducing} &   80.31 & 79.24 & 72.97 & 75.84 & 76.96 & 77.09 & 2.90 & 80.85 & 79.11 & 77.50 & 76.56 & 78.63 & 78.51 & 1.63 \\
        EoA \citep{arpit2021ensemble} & 82.30 & \underline{81.63} & 75.02 & \textbf{78.83} & \underline{79.32} & \underline{79.45} & 2.87 & \underline{82.88} & \underline{81.14} & \textbf{79.57} & \textbf{79.10} & \textbf{80.76} & \textbf{80.67} & 1.48 \\
        Mixstyle \citep{zhou2021domain} & 80.74 & 79.59 & 73.80 & 76.39 & 77.51 & 77.63 & 2.73 & 81.02 & 79.20 & 77.67 & 77.25 & 78.87 & 78.78 & 1.48 \\
        MLDG \citep{li2018learning} & 81.46 & 80.28 & 73.73 & 76.92 & 77.96 & 78.10 & 3.02 & 81.88 & 79.95 & 78.74 & 77.79 & 79.71 & 79.59 & 1.53 \\
        MMD \citep{li2018domain} &   81.37 & 80.63 & 73.82 & 77.10 & 78.12 & 78.23 &  3.01 & 81.93 & 80.28 & 78.71 & 77.85 & 79.81 & 79.69 & 1.56 \\
        CORAL \citep{sun2016deep} & \underline{82.66} & 81.36 & 74.70 & \underline{78.25} & 79.09 & 79.24 & 3.07 & 82.84 & 81.08 & \underline{79.49} & 78.82 & \underline{80.67} & \underline{80.56} & 1.55 \\
        StableNet \citep{zhang2021deep} & 81.52 & 80.36 & 76.17 & 77.29 & 78.85 & 78.84 & 2.18 & 82.56 & \textbf{82.21} & 78.35 & 77.46 & 80.12 & 80.15 & 2.27 \\
        FACT \citep{xu2021fourier} & 80.83 & 79.66 & 76.30 & 78.05 & 78.61 & 78.71 & \underline{1.71} & 81.37 & 79.39 & 78.06 & 78.58 & 79.37 & 79.35 & \textbf{1.26} \\
        JiGen \citep{carlucci2019domain} & 81.67 & 80.36 & \underline{76.54} & 78.17 & 79.08 & 79.18 & 1.98 & 81.64 & 79.84 & 78.14 & \underline{78.89} & 79.63 & 79.63 & \underline{1.31} \\
        GroupDRO \citep{sagawa2019distributionally} & 81.08 & 79.92 & 73.39 & 76.58 & 77.61 & 77.74 & 3.01 &   81.35 & 79.50 & 78.14 & 77.23 & 79.17 & 79.05 & 1.55 \\
        IRM \citep{arjovsky2019invariant} & 70.59 & 72.02 & 61.83 & 69.28 & 68.33 & 68.43 & 3.93 & 72.96 & 71.52 & 67.31 & 69.43 & 70.25 & 70.31 & 2.14 \\
        \midrule
        Oracle & 86.42 & 86.68 & 84.44 & 84.59 & 85.55 & 85.53 & 1.02 & 87.79 & 87.86 & 84.33 & 85.18 & 86.29 & 86.29 & 1.57 \\
        \bottomrule
    \end{tabular}}
    \tablelabel{tab:dg}
\end{table*}

\subsection{Benchmark for Flexible Domain Generalization}

Compared current DG setting where domains are aligned across categories, flexible combination of categories and domains in both training and test data can be more realistic and challenging \citep{zhang2021deep,shen2021towards}. In such cases,
the level of the distribution shifts varies in different classes, requiring a strong ability of generalization to tell common knowledge of categories from various domains. We present two settings, namely \textit{random} and \textit{compositional}. We randomly select two domains out of common domains as dominant ones, 12 out of the remaining domains as minor ones and the other 6 domains as test data for each category for the \textit{random} setting. There can be spurious correlations between domains and labels since a domain can be with class \textit{A} in training data and class \textit{B} in test data, while there can not be with class \textit{A} in both training and test data. For the \textit{compositional} setting, 4 domains are chosen as exclusive training domains and others as sharing domains. Then 2 domains are randomly selected from exclusive training domains as majority, 12 from sharing domains as minority and the remaining 4 in sharing domains for test. Thus there is no spurious correlations between dominant domains and labels. We select all images from dominant domains and 50 images from each minor domain for training and 50 images from each test domain for test.
Results are shown in \tableref{tab:ood}. Current SOTA algorithm outperforms ERM by a noticeable margin, yet the gap to Oracle remains significant. More splits, discussions and implementation details are in Appendix.

\begin{table}[t]
    \centering
    \caption{Results of the flexible DG setting on NICO$^{++}$.}
    \resizebox{\linewidth}{!}{
    \begin{tabular}{c|ccccccccccc|c}
        \toprule
        Method & Deepall & SWAD & MMLD & RSC & AdaClust & SagNet & EoA & MixStyle & StableNet & FACT & JiGen & Oracle \\
        \midrule
        Rand. & 74.19 & 75.62 & 73.25 & 75.20 & 73.39 & 72.79 & \underline{76.22} & 73.47 & \textbf{77.37} & 75.34 & 75.44 & 84.60 \\
        Comp. & 78.01 & 76.97 & 76.85 & 75.76 & 76.64 & 76.15 & \textbf{79.62} & 77.01 & 78.19 & \underline{79.39} & 78.77 & 86.18 \\
        Avg. & 76.10 & 76.30 & 75.05 & 75.48 & 75.02 & 74.47 & \textbf{77.92} & 75.24 & \underline{77.78} & 77.37 & 77.11 & 85.39 \\
        \bottomrule
    \end{tabular}
    }
    \tablelabel{tab:ood}
\end{table}

\begin{table}[t]
    \centering
    \caption{Standard deviation across epochs and seeds on different datasets.}
    \resizebox{\linewidth}{!}{
    \begin{tabular}{@{}c|ccc|ccc|ccc|ccc|ccc@{}}
    \toprule
              & \multicolumn{3}{c|}{PACS} & \multicolumn{3}{c|}{DomainNet}        & \multicolumn{3}{c|}{VLCS} & \multicolumn{3}{c|}{OfficeHome}       & \multicolumn{3}{c}{NICO$^{++}$}               \\ \midrule
    Method    & Epoch   & Seed   & Gap    & Epoch & Seed          & Gap           & Epoch   & Seed   & Gap    & Epoch & Seed          & Gap           & Epoch         & Seed          & Gap           \\ \midrule
    Deepall       & 0.96    & 0.82   & 2.66   & 0.61  & 0.57          & 0.46          & 0.83    & 0.58   & 3.59   & 0.77  & 0.59          & 0.81          & \textbf{0.22} & \textbf{0.10} & \textbf{0.39} \\
    SWAD      & 0.41    & 0.76   &    1.61    & 0.35  & 0.30          &     0.39          & 0.74    & 0.49   &  0.58  & 0.31  & 0.25       &    0.30              & \textbf{0.07} & \textbf{0.05} &  \textbf{0.06}             \\
    MMLD      & 1.68    & 2.02   &   3.25     & 1.03  & 0.50          &        0.85       & 2.33    & 1.12   &     3.97   & 1.25  & 0.47          &     0.56          & \textbf{0.25} & \textbf{0.10} &      \textbf{0.15}         \\
    RSC       & 0.76    & 0.81   &   0.93     & 0.55  & 0.35          &       0.56        & 1.02    & 0.61   &     0.80   & 0.85  & 0.37          &   0.89            & \textbf{0.18} & \textbf{0.05} &  \textbf{0.10}             \\
    AdaClust  & 1.06    & 1.74   & 1.54   & 0.98  & 0.41          & 0.72          & 1.32    & 1.79   & 1.34   & 1.36  & 1.30          & 0.28          & \textbf{0.22} & \textbf{0.04} & \textbf{0.13} \\
    SagNet    & 0.74    & 2.44   & 2.78   & 0.92  & \textbf{0.23} & 0.54          & 0.94    & 1.74   & 4.19   & 0.80  & 0.30          & \textbf{0.44} & \textbf{0.11} & 0.31          & 0.61          \\
    EoA       & 0.11    & 0.36   & 0.18   & 0.22  & 0.16          & \textbf{0.02} & 0.15    & 0.45   & 0.21   & 0.05  & 0.29          & 0.08          & \textbf{0.02} & \textbf{0.04} & 0.13          \\
    MixStyle  & 1.53    & 0.63   & 1.69   & 0.60  & 0.36          & 0.42          & 1.27    & 1.78   & 3.40   & 0.72  & 0.43          & 0.56          & \textbf{0.17} & \textbf{0.16} & \textbf{0.00} \\
    MLDG      & 0.82    & 1.02   & 1.24   & 0.53  & 0.25          & 0.55          & 1.15    & 1.01   & 4.14   & 1.03  & 0.09          & 0.23          & \textbf{0.10} & \textbf{0.08} & \textbf{0.12} \\
    MMD       & 1.13    & 2.39   & 0.66   & 0.82  & 0.24          & 0.50          & 1.98    & 1.32   & 3.72   & 0.61  & \textbf{0.02} & \textbf{1.34} & \textbf{0.11} & 0.11          & \textbf{0.16} \\
    CORAL     & 1.09    & 1.02   & 1.18   & 0.52  & 0.48          & 0.47          & 0.77    & 0.94   & 3.18   & 0.49  & 0.28          & 0.50          & \textbf{0.06} & \textbf{0.17} & \textbf{0.19} \\
    StableNet & 0.90    & 1.25   &   1.03     & 0.34  & 0.71          &       0.82        & 0.86    & 0.69   &   0.88     & 0.44  & 0.21          &      0.48         & \textbf{0.09} & \textbf{0.05} &   \textbf{0.09}            \\
    FACT      & 0.31    & 0.46   & 0.52   & 0.14  & \textbf{0.16} & \textbf{0.37} & 0.64    & 0.85   & 1.17   & 0.21  & 0.27          & 0.68          & \textbf{0.06} & 0.19          & 1.09          \\
    JiGen     & 0.33    & 1.15   & 0.70   & 0.16  & 0.18          & 0.39          & 0.51    & 0.67   & 1.30   & 0.20  & 0.69          & 0.25          & \textbf{0.05} & \textbf{0.09} & \textbf{0.10} \\
    GroupDRO  & 1.27    & 0.96   & 2.09   & 0.96  & 0.37          & 0.54          & 1.18    & 0.85   & 4.93   & 0.63  & 0.47          & 0.55          & \textbf{0.16} & \textbf{0.10} & \textbf{0.16} \\
    IRM       & 3.77    & 3.02   & 4.14   & 2.17  & 0.89          & 0.00          & 6.00    & 1.74   & 5.77   & 2.10  & 1.59          & 0.00          & \textbf{0.90} & \textbf{0.54} & \textbf{0.00} \\ \bottomrule
    \end{tabular}
    }
    \tablelabel{tab:variance}
\end{table}

\subsection{Test Variance and Model Selection}
Model selection (including the choice of hyperparameters, training checkpoints and architecture variants) effects DG evaluation considerably \citep{arpit2021ensemble,gulrajani2020search,shen2021towards}. The leak of knowledge of test data in training or model selection phase is criticized yet still usual in current algorithms \citep{gulrajani2020search,arpit2021ensemble}. 
This issue is exacerbated by the variance of test performance across random seeds, training iterations and other hyperparameters in that one can choose the best seed or the model from the best epoch under the guidance of released oracle validation set for a noticeable improvement. NICO$^{++}$ presents a feasible approach by reducing the test variance and thus decreasing the possible improvement by leveraging the leak. 

As shown in \sectionref{sec:evaluation}, the gap between the performance of a model on training and test data is bounded by the sum of covariant shift and concept shift between source and target domains. Intuitively, test variance on NICO$^{++}$ is lower than other current DG datasets given that NICO$^{++}$ guarantees a significantly lower concept shift. Strong concept shift between source domains introduces confusing mapping relations between input $\mathrm{X}$ and output $\mathrm{Y}$, embarrassing the convergence and enlarging the variance. 
Since most current deep models are optimized by stochastic gradient descent (SGD), the test accuracy is prone to jitter as the input sequence determined by random seeds varies.
Moreover, concept shift also grows the mismatch between the performance on validation data and test data, further widening the gap between target guided and source guided model selection. 


Empirically, we compare the test variance and the improvement of leveraging oracle knowledge on NICO$^{++}$ with other datasets across various seeds and training epochs in \tableref{tab:variance}. For the test variance across random seeds, we train 3 models for each method with 3 random seeds and calculate the test variance among them. For the test variance across epochs, we calculate the test variance of the models saved on the last 10 epochs for each random seed and show the mean value of 3 random seeds. NICO$^{++}$ shows a lower test variance compared with other datasets across both various random seeds and training epochs, indicating a more stable estimation of generalization ability robust to the choice of algorithm irrelevant hyperparameters. As a result, NICO$^{++}$ alleviates the oracle leaking issue by significantly squeezing the possible improvement space, leading to a fairer comparison for DG methods.

\section{Conclusion}
In this paper, we investigated the common grounds of advanced approaches for domain generalization in vision. To facilitate the progressive research, we proposed a context-extensive large-scale benchmark named NICO$^{++}$ along with more rational evaluation methods for comprehensively evaluating DG algorithms. Two metrics to quantify covariate shift and concept shift are proposed to evaluate DG datasets upon two novel generalization bounds. Extensive experiments showed the superiority of NICO$^{++}$ over current datasets and benchmarked DG algorithms comprehensively.

\clearpage

\bibliographystyle{plainnat}
\bibliography{egbib}

\begin{thebibliography}{97}
\providecommand{\natexlab}[1]{#1}
\providecommand{\url}[1]{\texttt{#1}}
\expandafter\ifx\csname urlstyle\endcsname\relax
  \providecommand{\doi}[1]{doi: #1}\else
  \providecommand{\doi}{doi: \begingroup \urlstyle{rm}\Url}\fi

\bibitem[Alcorn et~al.(2019)Alcorn, Li, Gong, Wang, Mai, Ku, and
  Nguyen]{alcorn2019strike}
Michael~A Alcorn, Qi~Li, Zhitao Gong, Chengfei Wang, Long Mai, Wei-Shinn Ku,
  and Anh Nguyen.
\newblock Strike (with) a pose: Neural networks are easily fooled by strange
  poses of familiar objects.
\newblock In \emph{Proceedings of the IEEE/CVF Conference on Computer Vision
  and Pattern Recognition}, pages 4845--4854, 2019.

\bibitem[Arjovsky et~al.(2019)Arjovsky, Bottou, Gulrajani, and
  Lopez-Paz]{arjovsky2019invariant}
Martin Arjovsky, L{\'e}on Bottou, Ishaan Gulrajani, and David Lopez-Paz.
\newblock Invariant risk minimization.
\newblock \emph{arXiv preprint arXiv:1907.02893}, 2019.

\bibitem[Arpit et~al.(2021)Arpit, Wang, Zhou, and Xiong]{arpit2021ensemble}
Devansh Arpit, Huan Wang, Yingbo Zhou, and Caiming Xiong.
\newblock Ensemble of averages: Improving model selection and boosting
  performance in domain generalization.
\newblock \emph{arXiv preprint arXiv:2110.10832}, 2021.

\bibitem[Bai et~al.(2020)Bai, Sun, Hong, Zhou, Ye, Ye, Chan, and
  Li]{bai2020decaug}
Haoyue Bai, Rui Sun, Lanqing Hong, Fengwei Zhou, Nanyang Ye, Han-Jia Ye,
  S-H~Gary Chan, and Zhenguo Li.
\newblock Decaug: Out-of-distribution generalization via decomposed feature
  representation and semantic augmentation.
\newblock \emph{arXiv preprint arXiv:2012.09382}, 2020.

\bibitem[Bandi et~al.(2018)Bandi, Geessink, Manson, Van~Dijk, Balkenhol,
  Hermsen, Bejnordi, Lee, Paeng, Zhong, et~al.]{bandi2018detection}
Peter Bandi, Oscar Geessink, Quirine Manson, Marcory Van~Dijk, Maschenka
  Balkenhol, Meyke Hermsen, Babak~Ehteshami Bejnordi, Byungjae Lee, Kyunghyun
  Paeng, Aoxiao Zhong, et~al.
\newblock From detection of individual metastases to classification of lymph
  node status at the patient level: the camelyon17 challenge.
\newblock \emph{IEEE transactions on medical imaging}, 38\penalty0
  (2):\penalty0 550--560, 2018.

\bibitem[Bartlett and Mendelson(2002)]{bartlett2002rademacher}
Peter~L Bartlett and Shahar Mendelson.
\newblock Rademacher and gaussian complexities: Risk bounds and structural
  results.
\newblock \emph{Journal of Machine search}, 3\penalty0 (Nov):\penalty0
  463--482, 2002.

\bibitem[Beery et~al.(2018)Beery, Van~Horn, and Perona]{beery2018recognition}
Sara Beery, Grant Van~Horn, and Pietro Perona.
\newblock Recognition in terra incognita.
\newblock In \emph{Proceedings of the European conference on computer vision
  (ECCV)}, pages 456--473, 2018.

\bibitem[Beery et~al.(2021)Beery, Agarwal, Cole, and
  Birodkar]{beery2021iwildcam}
Sara Beery, Arushi Agarwal, Elijah Cole, and Vighnesh Birodkar.
\newblock The iwildcam 2021 competition dataset.
\newblock \emph{arXiv preprint arXiv:2105.03494}, 2021.

\bibitem[Ben-David et~al.(2006)Ben-David, Blitzer, Crammer, and
  Pereira]{ben2006analysis}
Shai Ben-David, John Blitzer, Koby Crammer, and Fernando Pereira.
\newblock Analysis of representations for domain adaptation.
\newblock \emph{Advances in neural information processing systems}, 19, 2006.

\bibitem[Berman et~al.(2019)Berman, Buczak, Chavis, and
  Corbett]{berman2019survey}
Daniel~S Berman, Anna~L Buczak, Jeffrey~S Chavis, and Cherita~L Corbett.
\newblock A survey of deep learning methods for cyber security.
\newblock \emph{Information}, 10\penalty0 (4):\penalty0 122, 2019.

\bibitem[Blanchard et~al.(2011)Blanchard, Lee, and
  Scott]{blanchard2011generalizing}
Gilles Blanchard, Gyemin Lee, and Clayton Scott.
\newblock Generalizing from several related classification tasks to a new
  unlabeled sample.
\newblock \emph{NeurIPS}, 24:\penalty0 2178--2186, 2011.

\bibitem[Blanchard et~al.(2017)Blanchard, Deshmukh, Dogan, Lee, and
  Scott]{blanchard2017domain}
Gilles Blanchard, Aniket~Anand Deshmukh, Urun Dogan, Gyemin Lee, and Clayton
  Scott.
\newblock Domain generalization by marginal transfer learning.
\newblock \emph{arXiv preprint arXiv:1711.07910}, 2017.

\bibitem[Carlucci et~al.(2019)Carlucci, D'Innocente, Bucci, Caputo, and
  Tommasi]{carlucci2019domain}
Fabio~M Carlucci, Antonio D'Innocente, Silvia Bucci, Barbara Caputo, and
  Tatiana Tommasi.
\newblock Domain generalization by solving jigsaw puzzles.
\newblock In \emph{Proceedings of the IEEE/CVF Conference on Computer Vision
  and Pattern Recognition}, pages 2229--2238, 2019.

\bibitem[Castro et~al.(2020)Castro, Walker, and Glocker]{castro2020causality}
Daniel~C Castro, Ian Walker, and Ben Glocker.
\newblock Causality matters in medical imaging.
\newblock \emph{Nature Communications}, 11\penalty0 (1):\penalty0 1--10, 2020.

\bibitem[Cha et~al.(2021)Cha, Chun, Lee, Cho, Park, Lee, and Park]{cha2021swad}
Junbum Cha, Sanghyuk Chun, Kyungjae Lee, Han-Cheol Cho, Seunghyun Park, Yunsung
  Lee, and Sungrae Park.
\newblock Swad: Domain generalization by seeking flat minima.
\newblock \emph{Advances in Neural Information Processing Systems}, 34, 2021.

\bibitem[Christie et~al.(2018)Christie, Fendley, Wilson, and
  Mukherjee]{christie2018functional}
Gordon Christie, Neil Fendley, James Wilson, and Ryan Mukherjee.
\newblock Functional map of the world.
\newblock In \emph{Proceedings of the IEEE Conference on Computer Vision and
  Pattern Recognition}, pages 6172--6180, 2018.

\bibitem[Dai and Van~Gool(2018)]{dai2018dark}
Dengxin Dai and Luc Van~Gool.
\newblock Dark model adaptation: Semantic image segmentation from daytime to
  nighttime.
\newblock In \emph{2018 21st International Conference on Intelligent
  Transportation Systems (ITSC)}, pages 3819--3824. IEEE, 2018.

\bibitem[Deng et~al.(2009)Deng, Dong, Socher, Li, Li, and
  Fei-Fei]{deng2009imagenet}
Jia Deng, Wei Dong, Richard Socher, Li-Jia Li, Kai Li, and Li~Fei-Fei.
\newblock Imagenet: A large-scale hierarchical image database.
\newblock In \emph{2009 IEEE conference on computer vision and pattern
  recognition}, pages 248--255. Ieee, 2009.

\bibitem[Ding and Fu(2017)]{ding2017deep}
Zhengming Ding and Yun Fu.
\newblock Deep domain generalization with structured low-rank constraint.
\newblock \emph{IEEE Transactions on Image Processing}, 27\penalty0
  (1):\penalty0 304--313, 2017.

\bibitem[Duchi et~al.(2020)Duchi, Hashimoto, and
  Namkoong]{duchi2020distributionally}
John Duchi, Tatsunori Hashimoto, and Hongseok Namkoong.
\newblock Distributionally robust losses for latent covariate mixtures.
\newblock \emph{arXiv preprint arXiv:2007.13982}, 2020.

\bibitem[Everingham et~al.(2015)Everingham, Eslami, Van~Gool, Williams, Winn,
  and Zisserman]{everingham2015pascal}
Mark Everingham, SM~Eslami, Luc Van~Gool, Christopher~KI Williams, John Winn,
  and Andrew Zisserman.
\newblock The pascal visual object classes challenge: A retrospective.
\newblock \emph{International journal of computer vision}, 111\penalty0
  (1):\penalty0 98--136, 2015.

\bibitem[Fang et~al.(2013)Fang, Xu, and Rockmore]{fang2013unbiased}
Chen Fang, Ye~Xu, and Daniel~N Rockmore.
\newblock Unbiased metric learning: On the utilization of multiple datasets and
  web images for softening bias.
\newblock In \emph{Proceedings of the IEEE International Conference on Computer
  Vision}, pages 1657--1664, 2013.

\bibitem[Fang et~al.(2020)Fang, Lu, Niu, and Sugiyama]{fang2020rethinking}
Tongtong Fang, Nan Lu, Gang Niu, and Masashi Sugiyama.
\newblock Rethinking importance weighting for deep learning under distribution
  shift.
\newblock \emph{Advances in Neural Information Processing Systems},
  33:\penalty0 11996--12007, 2020.

\bibitem[Gan et~al.(2016)Gan, Yang, and Gong]{gan2016learning}
Chuang Gan, Tianbao Yang, and Boqing Gong.
\newblock Learning attributes equals multi-source domain generalization.
\newblock In \emph{CVPR}, pages 87--97, 2016.

\bibitem[Ganin and Lempitsky(2015)]{ganin2015unsupervised}
Yaroslav Ganin and Victor Lempitsky.
\newblock Unsupervised domain adaptation by backpropagation.
\newblock In \emph{International conference on machine learning}, pages
  1180--1189. PMLR, 2015.

\bibitem[Ganin et~al.(2016)Ganin, Ustinova, Ajakan, Germain, Larochelle,
  Laviolette, Marchand, and Lempitsky]{ganin2016domain}
Yaroslav Ganin, Evgeniya Ustinova, Hana Ajakan, Pascal Germain, Hugo
  Larochelle, Fran{\c{c}}ois Laviolette, Mario Marchand, and Victor Lempitsky.
\newblock Domain-adversarial training of neural networks.
\newblock \emph{The journal of machine learning research}, 17\penalty0
  (1):\penalty0 2096--2030, 2016.

\bibitem[Ghafoorian et~al.(2017)Ghafoorian, Mehrtash, Kapur, Karssemeijer,
  Marchiori, Pesteie, Guttmann, Leeuw, Tempany, Ginneken,
  et~al.]{ghafoorian2017transfer}
Mohsen Ghafoorian, Alireza Mehrtash, Tina Kapur, Nico Karssemeijer, Elena
  Marchiori, Mehran Pesteie, Charles~RG Guttmann, Frank-Erik~de Leeuw, Clare~M
  Tempany, Bram~van Ginneken, et~al.
\newblock Transfer learning for domain adaptation in mri: Application in brain
  lesion segmentation.
\newblock In \emph{International conference on medical image computing and
  computer-assisted intervention}, pages 516--524. Springer, 2017.

\bibitem[Ghifary et~al.(2015)Ghifary, Kleijn, Zhang, and
  Balduzzi]{ghifary2015domain}
Muhammad Ghifary, W~Bastiaan Kleijn, Mengjie Zhang, and David Balduzzi.
\newblock Domain generalization for object recognition with multi-task
  autoencoders.
\newblock In \emph{Proceedings of the IEEE international conference on computer
  vision}, pages 2551--2559, 2015.

\bibitem[Ghifary et~al.(2016)Ghifary, Balduzzi, Kleijn, and
  Zhang]{ghifary2016scatter}
Muhammad Ghifary, David Balduzzi, W~Bastiaan Kleijn, and Mengjie Zhang.
\newblock Scatter component analysis: A unified framework for domain adaptation
  and domain generalization.
\newblock \emph{IEEE TPAMI}, 39\penalty0 (7):\penalty0 1414--1430, 2016.

\bibitem[Grubinger et~al.(2015)Grubinger, Birlutiu, Sch{\"o}ner,
  Natschl{\"a}ger, and Heskes]{grubinger2015domain}
Thomas Grubinger, Adriana Birlutiu, Holger Sch{\"o}ner, Thomas Natschl{\"a}ger,
  and Tom Heskes.
\newblock Domain generalization based on transfer component analysis.
\newblock In \emph{International Work-Conference on Artificial Neural
  Networks}, pages 325--334. Springer, 2015.

\bibitem[Gulrajani and Lopez-Paz(2021)]{gulrajani2020search}
Ishaan Gulrajani and David Lopez-Paz.
\newblock In search of lost domain generalization.
\newblock In \emph{International Conference on Learning Representations}, 2021.

\bibitem[He et~al.(2016)He, Zhang, Ren, and Sun]{he2016deep}
Kaiming He, Xiangyu Zhang, Shaoqing Ren, and Jian Sun.
\newblock Deep residual learning for image recognition.
\newblock In \emph{Proceedings of the IEEE conference on computer vision and
  pattern recognition}, pages 770--778, 2016.

\bibitem[He et~al.(2021)He, Shen, and Cui]{he2021towards}
Yue He, Zheyan Shen, and Peng Cui.
\newblock Towards non-iid image classification: A dataset and baselines.
\newblock \emph{Pattern Recognition}, 110:\penalty0 107383, 2021.

\bibitem[Hendrycks and Dietterich(2019)]{hendrycks2019benchmarking}
Dan Hendrycks and Thomas Dietterich.
\newblock Benchmarking neural network robustness to common corruptions and
  perturbations.
\newblock \emph{arXiv preprint arXiv:1903.12261}, 2019.

\bibitem[Hendrycks et~al.(2021{\natexlab{a}})Hendrycks, Basart, Mu, Kadavath,
  Wang, Dorundo, Desai, Zhu, Parajuli, Guo, et~al.]{hendrycks2021many}
Dan Hendrycks, Steven Basart, Norman Mu, Saurav Kadavath, Frank Wang, Evan
  Dorundo, Rahul Desai, Tyler Zhu, Samyak Parajuli, Mike Guo, et~al.
\newblock The many faces of robustness: A critical analysis of
  out-of-distribution generalization.
\newblock In \emph{Proceedings of the IEEE/CVF International Conference on
  Computer Vision}, pages 8340--8349, 2021{\natexlab{a}}.

\bibitem[Hendrycks et~al.(2021{\natexlab{b}})Hendrycks, Zhao, Basart,
  Steinhardt, and Song]{hendrycks2021natural}
Dan Hendrycks, Kevin Zhao, Steven Basart, Jacob Steinhardt, and Dawn Song.
\newblock Natural adversarial examples.
\newblock In \emph{Proceedings of the IEEE/CVF Conference on Computer Vision
  and Pattern Recognition}, pages 15262--15271, 2021{\natexlab{b}}.

\bibitem[Hu et~al.(2020)Hu, Zhang, Chen, and Chan]{hu2020domain}
Shoubo Hu, Kun Zhang, Zhitang Chen, and Laiwan Chan.
\newblock Domain generalization via multidomain discriminant analysis.
\newblock In \emph{Uncertainty in Artificial Intelligence}, pages 292--302.
  PMLR, 2020.

\bibitem[Huang et~al.(2020)Huang, Wang, Xing, and Huang]{huang2020self}
Zeyi Huang, Haohan Wang, Eric~P Xing, and Dong Huang.
\newblock Self-challenging improves cross-domain generalization.
\newblock In \emph{European Conference on Computer Vision}, pages 124--140.
  Springer, 2020.

\bibitem[Ilse et~al.(2020)Ilse, Tomczak, Louizos, and Welling]{ilse2020diva}
Maximilian Ilse, Jakub~M Tomczak, Christos Louizos, and Max Welling.
\newblock Diva: Domain invariant variational autoencoders.
\newblock In \emph{Medical Imaging with Deep Learning}, pages 322--348. PMLR,
  2020.

\bibitem[Jin et~al.(2021)Jin, Lan, Zeng, and Chen]{jin2021style}
Xin Jin, Cuiling Lan, Wenjun Zeng, and Zhibo Chen.
\newblock Style normalization and restitution for domaingeneralization and
  adaptation.
\newblock \emph{arXiv preprint arXiv:2101.00588}, 2021.

\bibitem[Khirodkar et~al.(2019)Khirodkar, Yoo, and Kitani]{khirodkar2019domain}
Rawal Khirodkar, Donghyun Yoo, and Kris Kitani.
\newblock Domain randomization for scene-specific car detection and pose
  estimation.
\newblock In \emph{2019 IEEE Winter Conference on Applications of Computer
  Vision (WACV)}, pages 1932--1940. IEEE, 2019.

\bibitem[Kifer et~al.(2004)Kifer, Ben-David, and Gehrke]{kifer2004detecting}
Daniel Kifer, Shai Ben-David, and Johannes Gehrke.
\newblock Detecting change in data streams.
\newblock In \emph{VLDB}, volume~4, pages 180--191. Toronto, Canada, 2004.

\bibitem[Kipf and Welling(2016)]{kipf2016semi}
Thomas~N Kipf and Max Welling.
\newblock Semi-supervised classification with graph convolutional networks.
\newblock \emph{arXiv preprint arXiv:1609.02907}, 2016.

\bibitem[Koh et~al.(2021)Koh, Sagawa, Marklund, Xie, Zhang, Balsubramani, Hu,
  Yasunaga, Phillips, Gao, et~al.]{koh2021wilds}
Pang~Wei Koh, Shiori Sagawa, Henrik Marklund, Sang~Michael Xie, Marvin Zhang,
  Akshay Balsubramani, Weihua Hu, Michihiro Yasunaga, Richard~Lanas Phillips,
  Irena Gao, et~al.
\newblock Wilds: A benchmark of in-the-wild distribution shifts.
\newblock In \emph{International Conference on Machine Learning}, pages
  5637--5664. PMLR, 2021.

\bibitem[Levinson et~al.(2011)Levinson, Askeland, Becker, Dolson, Held, Kammel,
  Kolter, Langer, Pink, Pratt, et~al.]{levinson2011towards}
Jesse Levinson, Jake Askeland, Jan Becker, Jennifer Dolson, David Held, Soeren
  Kammel, J~Zico Kolter, Dirk Langer, Oliver Pink, Vaughan Pratt, et~al.
\newblock Towards fully autonomous driving: Systems and algorithms.
\newblock In \emph{2011 IEEE intelligent vehicles symposium (IV)}, pages
  163--168. IEEE, 2011.

\bibitem[Li et~al.(2017)Li, Yang, Song, and Hospedales]{li2017deeper}
Da~Li, Yongxin Yang, Yi-Zhe Song, and Timothy~M Hospedales.
\newblock Deeper, broader and artier domain generalization.
\newblock In \emph{Proceedings of the IEEE international conference on computer
  vision}, pages 5542--5550, 2017.

\bibitem[Li et~al.(2018{\natexlab{a}})Li, Yang, Song, and
  Hospedales]{li2018learning}
Da~Li, Yongxin Yang, Yi-Zhe Song, and Timothy~M Hospedales.
\newblock Learning to generalize: Meta-learning for domain generalization.
\newblock In \emph{Thirty-Second AAAI Conference on Artificial Intelligence},
  2018{\natexlab{a}}.

\bibitem[Li et~al.(2019)Li, Zhang, Yang, Liu, Song, and
  Hospedales]{li2019episodic}
Da~Li, Jianshu Zhang, Yongxin Yang, Cong Liu, Yi-Zhe Song, and Timothy~M
  Hospedales.
\newblock Episodic training for domain generalization.
\newblock In \emph{Proceedings of the IEEE/CVF International Conference on
  Computer Vision}, pages 1446--1455, 2019.

\bibitem[Li et~al.(2018{\natexlab{b}})Li, Pan, Wang, and Kot]{li2018domain}
Haoliang Li, Sinno~Jialin Pan, Shiqi Wang, and Alex~C Kot.
\newblock Domain generalization with adversarial feature learning.
\newblock In \emph{Proceedings of the IEEE conference on computer vision and
  pattern recognition}, pages 5400--5409, 2018{\natexlab{b}}.

\bibitem[Liao et~al.(2020)Liao, Huang, Li, Chen, and Li]{liao2020deep}
Yixiao Liao, Ruyi Huang, Jipu Li, Zhuyun Chen, and Weihua Li.
\newblock Deep semisupervised domain generalization network for rotary
  machinery fault diagnosis under variable speed.
\newblock \emph{IEEE Transactions on Instrumentation and Measurement},
  69\penalty0 (10):\penalty0 8064--8075, 2020.

\bibitem[Lin et~al.(2014)Lin, Maire, Belongie, Hays, Perona, Ramanan,
  Doll{\'a}r, and Zitnick]{lin2014microsoft}
Tsung-Yi Lin, Michael Maire, Serge Belongie, James Hays, Pietro Perona, Deva
  Ramanan, Piotr Doll{\'a}r, and C~Lawrence Zitnick.
\newblock Microsoft coco: Common objects in context.
\newblock In \emph{European conference on computer vision}, pages 740--755.
  Springer, 2014.

\bibitem[Mancini et~al.(2018)Mancini, Bulo, Caputo, and Ricci]{mancini2018best}
Massimiliano Mancini, Samuel~Rota Bulo, Barbara Caputo, and Elisa Ricci.
\newblock Best sources forward: domain generalization through source-specific
  nets.
\newblock In \emph{2018 25th IEEE international conference on image processing
  (ICIP)}, pages 1353--1357. IEEE, 2018.

\bibitem[Mansour et~al.(2009)Mansour, Mohri, and
  Rostamizadeh]{mansour2009domain}
Yishay Mansour, Mehryar Mohri, and Afshin Rostamizadeh.
\newblock Domain adaptation: Learning bounds and algorithms.
\newblock \emph{arXiv preprint arXiv:0902.3430}, 2009.

\bibitem[Matsuura and Harada(2020)]{matsuura2020domain}
Toshihiko Matsuura and Tatsuya Harada.
\newblock Domain generalization using a mixture of multiple latent domains.
\newblock In \emph{Proceedings of the AAAI Conference on Artificial
  Intelligence}, volume~34, pages 11749--11756, 2020.

\bibitem[Miotto et~al.(2018)Miotto, Wang, Wang, Jiang, and
  Dudley]{miotto2018deep}
Riccardo Miotto, Fei Wang, Shuang Wang, Xiaoqian Jiang, and Joel~T Dudley.
\newblock Deep learning for healthcare: review, opportunities and challenges.
\newblock \emph{Briefings in bioinformatics}, 19\penalty0 (6):\penalty0
  1236--1246, 2018.

\bibitem[Muandet et~al.(2013)Muandet, Balduzzi, and
  Sch{\"o}lkopf]{muandet2013domain}
Krikamol Muandet, David Balduzzi, and Bernhard Sch{\"o}lkopf.
\newblock Domain generalization via invariant feature representation.
\newblock In \emph{ICML}, pages 10--18. PMLR, 2013.

\bibitem[Nam and Kim(2018)]{nam2018batch}
Hyeonseob Nam and Hyo-Eun Kim.
\newblock Batch-instance normalization for adaptively style-invariant neural
  networks.
\newblock \emph{arXiv preprint arXiv:1805.07925}, 2018.

\bibitem[Nam et~al.(2021)Nam, Lee, Park, Yoon, and Yoo]{nam2021reducing}
Hyeonseob Nam, HyunJae Lee, Jongchan Park, Wonjun Yoon, and Donggeun Yoo.
\newblock Reducing domain gap by reducing style bias.
\newblock In \emph{Proceedings of the IEEE/CVF Conference on Computer Vision
  and Pattern Recognition}, pages 8690--8699, 2021.

\bibitem[Peng et~al.(2019)Peng, Bai, Xia, Huang, Saenko, and
  Wang]{peng2019moment}
Xingchao Peng, Qinxun Bai, Xide Xia, Zijun Huang, Kate Saenko, and Bo~Wang.
\newblock Moment matching for multi-source domain adaptation.
\newblock In \emph{Proceedings of the IEEE/CVF international conference on
  computer vision}, pages 1406--1415, 2019.

\bibitem[Peng et~al.(2018)Peng, Andrychowicz, Zaremba, and Abbeel]{peng2018sim}
Xue~Bin Peng, Marcin Andrychowicz, Wojciech Zaremba, and Pieter Abbeel.
\newblock Sim-to-real transfer of robotic control with dynamics randomization.
\newblock In \emph{2018 IEEE international conference on robotics and
  automation (ICRA)}, pages 3803--3810. IEEE, 2018.

\bibitem[Prakash et~al.(2019)Prakash, Boochoon, Brophy, Acuna, Cameracci,
  State, Shapira, and Birchfield]{prakash2019structured}
Aayush Prakash, Shaad Boochoon, Mark Brophy, David Acuna, Eric Cameracci,
  Gavriel State, Omer Shapira, and Stan Birchfield.
\newblock Structured domain randomization: Bridging the reality gap by
  context-aware synthetic data.
\newblock In \emph{2019 International Conference on Robotics and Automation
  (ICRA)}, pages 7249--7255. IEEE, 2019.

\bibitem[Qui{\~n}onero-Candela et~al.(2008)Qui{\~n}onero-Candela, Sugiyama,
  Schwaighofer, and Lawrence]{quinonero2008dataset}
Joaquin Qui{\~n}onero-Candela, Masashi Sugiyama, Anton Schwaighofer, and Neil~D
  Lawrence.
\newblock \emph{Dataset shift in machine learning}.
\newblock Mit Press, 2008.

\bibitem[Ruan et~al.(2021)Ruan, Dubois, and Maddison]{ruan2021optimal}
Yangjun Ruan, Yann Dubois, and Chris~J Maddison.
\newblock Optimal representations for covariate shift.
\newblock \emph{arXiv preprint arXiv:2201.00057}, 2021.

\bibitem[Ryu et~al.(2019)Ryu, Kwon, Yang, and Lim]{ryu2019generalized}
Jongbin Ryu, Gitaek Kwon, Ming-Hsuan Yang, and Jongwoo Lim.
\newblock Generalized convolutional forest networks for domain generalization
  and visual recognition.
\newblock In \emph{ICLR}, 2019.

\bibitem[Sagawa et~al.(2019)Sagawa, Koh, Hashimoto, and
  Liang]{sagawa2019distributionally}
Shiori Sagawa, Pang~Wei Koh, Tatsunori~B Hashimoto, and Percy Liang.
\newblock Distributionally robust neural networks for group shifts: On the
  importance of regularization for worst-case generalization.
\newblock \emph{arXiv preprint arXiv:1911.08731}, 2019.

\bibitem[Segu et~al.(2020)Segu, Tonioni, and Tombari]{segu2020batch}
Mattia Segu, Alessio Tonioni, and Federico Tombari.
\newblock Batch normalization embeddings for deep domain generalization.
\newblock \emph{arXiv preprint arXiv:2011.12672}, 2020.

\bibitem[Sener et~al.(2016)Sener, Song, Saxena, and
  Savarese]{sener2016learning}
Ozan Sener, Hyun~Oh Song, Ashutosh Saxena, and Silvio Savarese.
\newblock Learning transferrable representations for unsupervised domain
  adaptation.
\newblock \emph{Advances in neural information processing systems}, 29, 2016.

\bibitem[Shankar et~al.(2018)Shankar, Piratla, Chakrabarti, Chaudhuri, Jyothi,
  and Sarawagi]{shankar2018generalizing}
Shiv Shankar, Vihari Piratla, Soumen Chakrabarti, Siddhartha Chaudhuri, Preethi
  Jyothi, and Sunita Sarawagi.
\newblock Generalizing across domains via cross-gradient training.
\newblock \emph{arXiv preprint arXiv:1804.10745}, 2018.

\bibitem[Shen et~al.(2021)Shen, Liu, He, Zhang, Xu, Yu, and
  Cui]{shen2021towards}
Zheyan Shen, Jiashuo Liu, Yue He, Xingxuan Zhang, Renzhe Xu, Han Yu, and Peng
  Cui.
\newblock Towards out-of-distribution generalization: A survey.
\newblock \emph{arXiv preprint arXiv:2108.13624}, 2021.

\bibitem[Simonyan and Zisserman(2014)]{simonyan2014very}
Karen Simonyan and Andrew Zisserman.
\newblock Very deep convolutional networks for large-scale image recognition.
\newblock \emph{arXiv preprint arXiv:1409.1556}, 2014.

\bibitem[Sugiyama et~al.(2007{\natexlab{a}})Sugiyama, Krauledat, and
  M{\"u}ller]{sugiyama2007covariate}
Masashi Sugiyama, Matthias Krauledat, and Klaus-Robert M{\"u}ller.
\newblock Covariate shift adaptation by importance weighted cross validation.
\newblock \emph{Journal of Machine Learning Research}, 8\penalty0 (5),
  2007{\natexlab{a}}.

\bibitem[Sugiyama et~al.(2007{\natexlab{b}})Sugiyama, Nakajima, Kashima,
  Buenau, and Kawanabe]{sugiyama2007direct}
Masashi Sugiyama, Shinichi Nakajima, Hisashi Kashima, Paul Buenau, and Motoaki
  Kawanabe.
\newblock Direct importance estimation with model selection and its application
  to covariate shift adaptation.
\newblock \emph{Advances in neural information processing systems}, 20,
  2007{\natexlab{b}}.

\bibitem[Sun and Saenko(2016)]{sun2016deep}
Baochen Sun and Kate Saenko.
\newblock Deep coral: Correlation alignment for deep domain adaptation.
\newblock In \emph{European conference on computer vision}, pages 443--450.
  Springer, 2016.

\bibitem[Tahmoresnezhad and Hashemi(2017)]{tahmoresnezhad2017visual}
Jafar Tahmoresnezhad and Sattar Hashemi.
\newblock Visual domain adaptation via transfer feature learning.
\newblock \emph{Knowledge and information systems}, 50\penalty0 (2):\penalty0
  585--605, 2017.

\bibitem[Thomas et~al.(2021)Thomas, Mahajan, Pentland, and
  Dubey]{thomas2021adaptive}
Xavier Thomas, Dhruv Mahajan, Alex Pentland, and Abhimanyu Dubey.
\newblock Adaptive methods for aggregated domain generalization.
\newblock \emph{arXiv preprint arXiv:2112.04766}, 2021.

\bibitem[Tobin et~al.(2017)Tobin, Fong, Ray, Schneider, Zaremba, and
  Abbeel]{tobin2017domain}
Josh Tobin, Rachel Fong, Alex Ray, Jonas Schneider, Wojciech Zaremba, and
  Pieter Abbeel.
\newblock Domain randomization for transferring deep neural networks from
  simulation to the real world.
\newblock In \emph{2017 IEEE/RSJ international conference on intelligent robots
  and systems (IROS)}, pages 23--30. IEEE, 2017.

\bibitem[Tremblay et~al.(2018)Tremblay, Prakash, Acuna, Brophy, Jampani, Anil,
  To, Cameracci, Boochoon, and Birchfield]{tremblay2018training}
Jonathan Tremblay, Aayush Prakash, David Acuna, Mark Brophy, Varun Jampani, Cem
  Anil, Thang To, Eric Cameracci, Shaad Boochoon, and Stan Birchfield.
\newblock Training deep networks with synthetic data: Bridging the reality gap
  by domain randomization.
\newblock In \emph{CVPR workshops}, pages 969--977, 2018.

\bibitem[Venkateswara et~al.(2017)Venkateswara, Eusebio, Chakraborty, and
  Panchanathan]{venkateswara2017deep}
Hemanth Venkateswara, Jose Eusebio, Shayok Chakraborty, and Sethuraman
  Panchanathan.
\newblock Deep hashing network for unsupervised domain adaptation.
\newblock In \emph{Proceedings of the IEEE conference on computer vision and
  pattern recognition}, pages 5018--5027, 2017.

\bibitem[Volpi et~al.(2018)Volpi, Namkoong, Sener, Duchi, Murino, and
  Savarese]{volpi2018generalizing}
Riccardo Volpi, Hongseok Namkoong, Ozan Sener, John~C Duchi, Vittorio Murino,
  and Silvio Savarese.
\newblock Generalizing to unseen domains via adversarial data augmentation.
\newblock \emph{Advances in neural information processing systems}, 31, 2018.

\bibitem[Wang et~al.(2021)Wang, Lan, Liu, Ouyang, Zeng, and
  Qin]{wang2021generalizing}
Jindong Wang, Cuiling Lan, Chang Liu, Yidong Ouyang, Wenjun Zeng, and Tao Qin.
\newblock Generalizing to unseen domains: A survey on domain generalization.
\newblock \emph{arXiv preprint arXiv:2103.03097}, 2021.

\bibitem[Wang et~al.(2020)Wang, Yu, Li, Yang, Fu, and Heng]{wang2020dofe}
Shujun Wang, Lequan Yu, Kang Li, Xin Yang, Chi-Wing Fu, and Pheng-Ann Heng.
\newblock Dofe: Domain-oriented feature embedding for generalizable fundus
  image segmentation on unseen datasets.
\newblock \emph{IEEE TMI}, 39\penalty0 (12):\penalty0 4237--4248, 2020.

\bibitem[Xu et~al.(2021{\natexlab{a}})Xu, Ebner, Yarmohammadi, White,
  Van~Durme, and Murray]{xu2021gradual}
Haoran Xu, Seth Ebner, Mahsa Yarmohammadi, Aaron~Steven White, Benjamin
  Van~Durme, and Kenton Murray.
\newblock Gradual fine-tuning for low-resource domain adaptation.
\newblock \emph{arXiv preprint arXiv:2103.02205}, 2021{\natexlab{a}}.

\bibitem[Xu et~al.(2020)Xu, Zhang, Li, Du, Kawarabayashi, and
  Jegelka]{xu2020neural}
Keyulu Xu, Mozhi Zhang, Jingling Li, Simon~S Du, Ken-ichi Kawarabayashi, and
  Stefanie Jegelka.
\newblock How neural networks extrapolate: From feedforward to graph neural
  networks.
\newblock \emph{arXiv preprint arXiv:2009.11848}, 2020.

\bibitem[Xu et~al.(2021{\natexlab{b}})Xu, Zhang, Zhang, Wang, and
  Tian]{xu2021fourier}
Qinwei Xu, Ruipeng Zhang, Ya~Zhang, Yanfeng Wang, and Qi~Tian.
\newblock A fourier-based framework for domain generalization.
\newblock In \emph{Proceedings of the IEEE/CVF Conference on Computer Vision
  and Pattern Recognition}, pages 14383--14392, 2021{\natexlab{b}}.

\bibitem[Ye et~al.(2021)Ye, Li, Hong, Bai, Chen, Zhou, and Li]{ye2021ood}
Nanyang Ye, Kaican Li, Lanqing Hong, Haoyue Bai, Yiting Chen, Fengwei Zhou, and
  Zhenguo Li.
\newblock Ood-bench: Benchmarking and understanding out-of-distribution
  generalization datasets and algorithms.
\newblock \emph{arXiv preprint arXiv:2106.03721}, 2021.

\bibitem[Young et~al.(2018)Young, Hazarika, Poria, and
  Cambria]{young2018recent}
Tom Young, Devamanyu Hazarika, Soujanya Poria, and Erik Cambria.
\newblock Recent trends in deep learning based natural language processing.
\newblock \emph{ieee Computational intelligenCe magazine}, 13\penalty0
  (3):\penalty0 55--75, 2018.

\bibitem[Yue et~al.(2019)Yue, Zhang, Zhao, Sangiovanni-Vincentelli, Keutzer,
  and Gong]{yue2019domain}
Xiangyu Yue, Yang Zhang, Sicheng Zhao, Alberto Sangiovanni-Vincentelli, Kurt
  Keutzer, and Boqing Gong.
\newblock Domain randomization and pyramid consistency: Simulation-to-real
  generalization without accessing target domain data.
\newblock In \emph{ICCV}, pages 2100--2110, 2019.

\bibitem[Zhang et~al.(2016)Zhang, Zuo, and Zhang]{zhang2016lsdt}
Lei Zhang, Wangmeng Zuo, and David Zhang.
\newblock Lsdt: Latent sparse domain transfer learning for visual adaptation.
\newblock \emph{IEEE Transactions on Image Processing}, 25\penalty0
  (3):\penalty0 1177--1191, 2016.

\bibitem[Zhang et~al.(2021{\natexlab{a}})Zhang, Cui, Xu, Zhou, He, and
  Shen]{zhang2021deep}
Xingxuan Zhang, Peng Cui, Renzhe Xu, Linjun Zhou, Yue He, and Zheyan Shen.
\newblock Deep stable learning for out-of-distribution generalization.
\newblock In \emph{Proceedings of the IEEE/CVF Conference on Computer Vision
  and Pattern Recognition}, pages 5372--5382, 2021{\natexlab{a}}.

\bibitem[Zhang et~al.(2021{\natexlab{b}})Zhang, Zhou, Xu, Cui, Shen, and
  Liu]{zhang2021domain}
Xingxuan Zhang, Linjun Zhou, Renzhe Xu, Peng Cui, Zheyan Shen, and Haoxin Liu.
\newblock Domain-irrelevant representation learning for unsupervised domain
  generalization.
\newblock \emph{arXiv preprint arXiv:2107.06219}, 2021{\natexlab{b}}.

\bibitem[Zhang et~al.(2020)Zhang, Deng, Tang, Zhang, and
  Jia]{zhang2020unsupervised}
Yabin Zhang, Bin Deng, Hui Tang, Lei Zhang, and Kui Jia.
\newblock Unsupervised multi-class domain adaptation: Theory, algorithms, and
  practice.
\newblock \emph{IEEE Transactions on Pattern Analysis and Machine
  Intelligence}, 2020.

\bibitem[Zhang et~al.(2019)Zhang, Liu, Long, and Jordan]{zhang2019bridging}
Yuchen Zhang, Tianle Liu, Mingsheng Long, and Michael Jordan.
\newblock Bridging theory and algorithm for domain adaptation.
\newblock In \emph{International Conference on Machine Learning}, pages
  7404--7413. PMLR, 2019.

\bibitem[Zhao et~al.(2019)Zhao, Des~Combes, Zhang, and
  Gordon]{zhao2019learning}
Han Zhao, Remi~Tachet Des~Combes, Kun Zhang, and Geoffrey Gordon.
\newblock On learning invariant representations for domain adaptation.
\newblock In \emph{International Conference on Machine Learning}, pages
  7523--7532. PMLR, 2019.

\bibitem[Zhao et~al.(2020)Zhao, Gong, Liu, Fu, and Tao]{zhao2020domain}
Shanshan Zhao, Mingming Gong, Tongliang Liu, Huan Fu, and Dacheng Tao.
\newblock Domain generalization via entropy regularization.
\newblock \emph{Advances in Neural Information Processing Systems},
  33:\penalty0 16096--16107, 2020.

\bibitem[Zhou et~al.(2020)Zhou, Yang, Hospedales, and Xiang]{zhou2020deep}
Kaiyang Zhou, Yongxin Yang, Timothy Hospedales, and Tao Xiang.
\newblock Deep domain-adversarial image generation for domain generalisation.
\newblock In \emph{Proceedings of the AAAI Conference on Artificial
  Intelligence}, volume~34, pages 13025--13032, 2020.

\bibitem[Zhou et~al.(2021{\natexlab{a}})Zhou, Liu, Qiao, Xiang, and
  Change~Loy]{zhou2021domain1}
Kaiyang Zhou, Ziwei Liu, Yu~Qiao, Tao Xiang, and Chen Change~Loy.
\newblock Domain generalization: A survey.
\newblock \emph{arXiv e-prints}, pages arXiv--2103, 2021{\natexlab{a}}.

\bibitem[Zhou et~al.(2021{\natexlab{b}})Zhou, Yang, Qiao, and
  Xiang]{zhou2021domain}
Kaiyang Zhou, Yongxin Yang, Yu~Qiao, and Tao Xiang.
\newblock Domain generalization with mixstyle.
\newblock \emph{arXiv preprint arXiv:2104.02008}, 2021{\natexlab{b}}.

\end{thebibliography}
\clearpage

\appendix

\section{More Theoretical Results and Discussions}
\subsection{Empirical version of \theoremref{thrm:bound-population} and \theoremref{thrm:bound-population-minus}}
Let $\hat{\calD}_\source$ and $\hat{\calD}_\target$ be the empirical training/testing distribution and $\hat{\varepsilon}_\source$ be the empirical loss with finite samples. We first introduce the empirical Rademacher complexity.

\begin{definition} [Empirical Rademacher Complexity \citep{bartlett2002rademacher}] \definitionlabel{defn:rademacher}
    Let $\calG$ be a set of real-valued functions defined over $\calX$. Given a sample $S \in \calX^n$, the empirical Rademacher Complexity of $\calG$ is defines as follows:
    \begin{equation}
        \hat{\rad}_S(\calG) = \frac{2}{n} \bbE_{\boldsymbol{\sigma}}\left[\left.\sup_{g \in \calG}\left|\sum_{i=1}^n\sigma_ig\left(x^{(i)}\right)\right|\,\,\right| \,\, S = \left(x^{(1)}, x^{(2)}, \dots, x^{(n)}\right)\right].
    \end{equation}
    Here $\boldsymbol{\sigma} = \{\sigma_i\}_{i=1}^n$ and $\sigma_i$ are \textit{i.i.d.} uniform random variables taking values in $\{+1, -1\}$.
\end{definition}

With \definitionref{defn:rademacher}, we can provide data-dependent bounds from empirical samples for \theoremref{thrm:bound-population} and \theoremref{thrm:bound-population-minus}.

\begin{theorem} \theoremlabel{thrm:bound-empirical}
    Suppose the loss function $\ell$ is symmetric, bounded by $M > 0$, and obeys the triangle inequality. Suppose $f_{\source}, f_{\target} \in \calH$. Then for any $\delta > 0$, with probability at least $1 - \delta$ over samples $S_{\source}$ of size $n_{\source}$ and $S_{\target}$ of size $n_{\target}$, the following inequality holds for all $h \in \calH$,
    \begin{equation}
        \begin{aligned}
            \varepsilon_{\target}(h) \le & \, \hat{\varepsilon}_{\source}(h) + \calM_{\concept}\left(\hat{\calD}_{\source}, \hat{\calD}_{\target}; \calH, \ell\right) + \calM_{\concept}^{\min}\left(\calD_\source, \calD_\target, f_\source, f_\target; \ell\right) \\
            & + \hat{\rad}_{S_\source}(\calL_{\calH}) + \hat{\rad}_{S_\target}(\calL_{\calH}) + \hat{\rad}_{S_\source}(\ell \circ \calH) + O\left(\sqrt{\frac{\log (1 / \delta)}{n_{\source}}}+\sqrt{\frac{\log (1 / \delta)}{n_{\target}}}\right).
        \end{aligned}
    \end{equation}
    Here $\calL_{\calH} \triangleq \left\{x \mapsto \ell(h(x), h'(x)): h, h' \in \calH\right\}$ and $\ell \circ \calH \triangleq \{(x, y) \mapsto \ell(h(x), y): h \in \calH\}$.
\end{theorem}

\begin{theorem} \theoremlabel{thrm:bound-empirical-minus}
    Suppose the loss function $\ell$ is symmetric, bounded by $M > 0$, and obeys the triangle inequality. Suppose $f_{\source}, f_{\target} \in \calH$. Then for any $\delta > 0$, with probability at least $1 - \delta$ over samples $S_{\source}$ of size $n_{\source}$ and $S_{\target}$ of size $n_{\target}$, the following inequality holds for all $h \in \calH$,
    \begin{equation}
        \begin{aligned}
            \varepsilon_{\target}(h) \ge & \, \calM_{\concept}^{\max}\left(\calD_\source, \calD_\target, f_\source, f_\target; \ell\right) - \calM_{\concept}\left(\hat{\calD}_{\source}, \hat{\calD}_{\target}; \calH, \ell\right) - \hat{\varepsilon}_{\source}(h) \\
            & - \hat{\rad}_{S_\source}(\calL_{\calH}) - \hat{\rad}_{S_\target}(\calL_{\calH}) - \hat{\rad}_{S_\source}(\ell \circ \calH) - O\left(\sqrt{\frac{\log (1 / \delta)}{n_{\source}}}+\sqrt{\frac{\log (1 / \delta)}{n_{\target}}}\right).
        \end{aligned}
    \end{equation}
    Here $\calL_{\calH} \triangleq \left\{x \mapsto \ell(h(x), h'(x)): h, h' \in \calH\right\}$ and $\ell \circ \calH \triangleq \{(x, y) \mapsto \ell(h(x), y): h \in \calH\}$.
\end{theorem}

\theoremref{thrm:bound-empirical} and \theoremref{thrm:bound-empirical-minus} quantify the effect of finite sample size to the bounds given by \theoremref{thrm:bound-population} and \theoremref{thrm:bound-population-minus}. Generally the bounds are tighter as the sample size increases and when the sample size tends towards infinity the bounds are identical to those given in \theoremref{thrm:bound-population} and \theoremref{thrm:bound-population-minus}, which meets the intuition.

\subsection{An Intuitively Explanation of Proposed Metrics}

Intuitively, the covariate shift in a dataset, which indicates how diversity of images across domains, should be strongly correlated with the distinction of domains. So that we connect the proposed metrics with the classification on domains.

As shown in \citep{mansour2009domain}, the discrepancy distance is a general formulation of the $d_\calA$-distance proposed in \citep{ben2006analysis}, which is defined as follows.
\begin{definition} [$d_\calA$-Distance \citep{kifer2004detecting}]
    Let $\calA$ be a set of subsets of $\calX$. The $d_\calA$-distance between two distributions $\calD_\source$ and $\calD_\target$ (with probability density $p_\source$ and $p_\target$ respectively) over $\calX$ is defined as
    \begin{equation}
        d_\calA(\calD_\source, \calD_\target) \triangleq \sup_{a \in \calA} |p_\source(a) - p_\target(a)|.
    \end{equation}
\end{definition}
According to \citep{mansour2009domain}, when $\calH = \{f: \calX \rightarrow \{0, 1\}\}$ is a set of binary classification functions and $\ell$ is set as the 0-1 classification loss, the discrepancy distance $\dd(\calD_\source, \calD_\target; \calH, \ell)$ coincides with the $d_\calA$-distance with $\calA = \{\{x: h(x) = 1\}: \forall h \in \tilde{\calH}\}$ and $\tilde{\calH} = \calH \Delta \calH \triangleq \left\{|h'-h|: h, h' \in \calH\right\}$. Furthermore,
\begin{equation}
\begin{aligned}
    d_\calA(\calD_\source, \calD_\target)  & = \sup_{a \in \calA} |p_\source(a) - p_\target(a)| = \sup_{h \in \tilde{\calH}} \left|\bbE_{x \in \calD_\source}[h(x)] - \bbE_{x \in \calD_\target}[h(x)]\right| \\
    & = 2\sup_{h \in \tilde{\calH}} \underbrace{\frac{1}{2}\left(\bbE_{x \in \calD_\source}[h(x)] + \bbE_{x \in \calD_\target}[1 - h(x)]\right)}_{\text{prediction accuracy on domains}} - 1 \\
\end{aligned}
\end{equation}
The last equality is due to the property that $h \in \tilde{\calH} \Longrightarrow 1 - h \in \tilde{\calH}$. Therefore, the $d_\calA$-distance is in terms of the optimal accuracy when classifying domains with functions in $\tilde{\calH}$.

As a result, the proposed covariate shift metric is strongly connected to a binary classification on training/test domains. If we split a dataset into training and test subsets according to domains, the more distinguishable these subsets are, the stronger covariate shift is within the dataset.

\subsection{Comparison between the Proposed Metrics and Kullback-Leibler Divergence}
We slightly abuse notations here to use $\calD_\source$ and $\calD_\target$ to denote the training distribution and testing distribution on $\calX \times \calY$ with probability density function $p_\source(x, y)$ and $p_\target(x, y)$ respectively. In addition, we use $\calD_\source^{\calX}$ and $\calD_\target^{\calX}$ to denote the marginal distribution of $\calD_\source$ and $\calD_\target$ on $\calX$.
\begin{equation} \equationlabel{eq:kl}
    \begin{aligned}
        & \kl\left(\calD_\source \| \calD_\target\right) \\
        = & \int_\calX\int_\calY p_\source(x, y)\log\frac{p_\source(x, y)}{p_\target(x, y)} \mathrm{d}x\mathrm{d}y \\
        = & \int_\calX\int_\calY p_\source(x, y)\log\frac{p_\source(y|x)}{p_\target(y|x)} \mathrm{d}x\mathrm{d}y + \int_\calX\int_\calY p_\source(x, y)\log\frac{p_\source(x)}{p_\target(x)} \mathrm{d}x\mathrm{d}y \\
        = & \int_\calX p_\source(x) \int_\calY p_\source(y|x) \log\frac{p_\source(y|x)}{p_\target(y|x)} \mathrm{d}y\mathrm{d}x + \int_\calX p_\source(x)\log\frac{p_\source(x)}{p_\target(x)} \mathrm{d}x \\
        = & \, \underbrace{\bbE_{x \sim \calD_\source^{\calX}}\left[\kl\left(p_\source(y|x)\| p_\target(y|x)\right)\right]}_{\text{Concept shift}} + \underbrace{\kl\left(\calD_\source^{\calX} \|  \calD_\target^{\calX}\right)}_{\text{Covariate shift}}.
    \end{aligned}
\end{equation}

Similar to our proposed metric $\calM_\covariate$ and $\calM_\concept$, the KL divergence between the training domain and testing domain could be divided into two parts, which measures the concept shift and covariate shift, respectively. However, compared to the RHS of \equationref{eq:kl}, our proposed metrics could bring two advantages. Firstly, our proposed metrics are easier to approximate with finite samples in practice (as shown in Section 4.3 in the main paper and A.1 and A.2 in Appendix) while the estimation of KL divergence is challenging \citep{wang2021generalizing,zhao2020domain}. Secondly, our proposed metrics have close connections with the error of models (as shown in \theoremref{thrm:bound-population} and \theoremref{thrm:bound-population-minus}), so that they are more befitting the evaluation of DG datasets for benchmarking DG algorithms. As a result, we adopt $\calM_\covariate$ and $\calM_\concept$ defined in the main body as the measures of covariate shift and concept shift.

\subsection{Comparison with Other Metrics}

Recently, some work tried to identify and measure distribution shifts in DG datasets \citep{bai2020decaug,ye2021ood}. Specifically, \citep{ye2021ood} proposed to group current DG datasets to two clusters, namely ones dominated by diversity shift and ones dominated by correlation shift. It assumes that 1) both training and test domains share the same labeling rule (\textit{i.e.}, $f_{\source} = f_{\target}$) and 2) there is no label shift across domains (\textit{i.e.}, $p_{\source}(Y) = p_{\target}(Y)$), which are unrestricted in our theorems. Especially, the metric \textit{concept shift} is proposed to measure how strong the labeling rule shifts between training and test domains. Moreover, the circumscription and calculation of diversity shift and correlation shift in \citep{ye2021ood} is based on variables related to $X$ but irrelevant to $Y$, and they require to be identified and split from $X$ initially, which can be challenging and even unsolvable \citep{shen2021towards,zhang2021deep}. While our metrics are defined according to $X$ itself and straightforward to estimate. 

\section{Important Lemmas and Omitted Proofs}

\subsection{Important lemmas}
\begin{lemma} [Rademacher Bound \citep{mansour2009domain}] \lemmalabel{lemma:rademacher}
    Let $\calG$ be a class of functions mapping $\calZ = \calX \times \calY$ to $[0, M]$ and $S = (z_1, z_2, \dots, z_n)$ a finite sample drawn \textit{i.i.d.} according to a distribution $\calD$. Then for any $\delta > 0$, with probability at least $1 - \delta$ over samples $S$ of size $n$, the following inequality holds for all $g \in \calG$,
    \begin{equation*}
        \calL_{\calD}(g) \le \hat{\calL}_{\calD}(g) + \hat{\rad}_S(\calG) + 3M\sqrt{\frac{\log (2 / \delta)}{2n}}.
    \end{equation*}
\end{lemma}

\begin{lemma}[Generalization bound for discrepancy distance \citep{mansour2009domain}]
\lemmalabel{lemma:generalization-dd}
    Assume that the loss function $\ell$ is bounded by $M > 0$. Let $\calD$ be a distribution over $\calX$ and let $\hat{\calD}$ denote the corresponding empirical distribution for a sample $S = (x_1, x_2, \dots, x_n)$. Then for any $\delta > 0$, with probability at least $1 - \delta$ over sample $S$ of size $n$ drawn according to $P$,
    \begin{equation*}
        \dd\left(\calD, \hat{\calD}; \calH, \ell\right) \le \hat{\rad}_S(\calL_{\calH}) + 3M\sqrt{\frac{\log (2 / \delta)}{2n}}.
    \end{equation*}
    Here $\calL_{\calH} \triangleq \left\{x \mapsto \ell(h(x), h'(x)): h, h' \in \calH\right\}$.
\end{lemma}

\subsection{Proof of \propositionref{prop:dd}}
\begin{proof}
    First, we know that
    $$
    \begin{aligned}
        & \, \dd\left(\calD_1, \calD_2; \calH, \ell\right) \\
        = & \,\sup_{h_1, h_2 \in \calH}\left|\calL_{\calD_1}(h_1, h_2) - \calL_{\calD_2}(h_1, h_2)\right| \\
        = & \, \max\left\{\sup_{h_1, h_2 \in \calH}\calL_{\calD_1}(h_1, h_2) - \calL_{\calD_2}(h_1, h_2), \sup_{h_1, h_2 \in \calH}\calL_{\calD_2}(h_1, h_2) - \calL_{\calD_1}(h_1, h_2)\right\}.
    \end{aligned}
    $$
    When $\calH$ is the set of all possible functions,
    $$
    \begin{aligned}
        & \, \sup_{h_1, h_2 \in \calH}\calL_{\calD_1}(h_1, h_2) - \calL_{\calD_2}(h_1, h_2) \\
        = & \, \sup_{h_1, h_2 \in \calH} \int_{\calX}\ell(h_1(x), h_2(x))(p_1(x) - p_2(x))\mathrm{d}x \\
        = & \,\int_{\calX} \left(\sup_{y_1, y_2 \in \calY}\ell(y_1, y_2)(p_1(x) - p_2(x))\right)\mathrm{d}x \\
        = & \, M \int_{\calX}\max\left\{p_1(x) - p_2(x), 0\right\} \mathrm{d} x \\
        = & \, \frac{M}{2}\int_{\calX}|p_1(x) - p_2(x)|\mathrm{d}x = \frac{M}{2}\ell_1(\calD_1, \calD_2).
    \end{aligned}
    $$
    Similarly, we can get that $\sup_{h_1, h_2 \in \calH}\calL_{\calD_2}(h_1, h_2) - \calL_{\calD_1}(h_1, h_2) = \frac{M}{2}\ell_1(\calD_1, \calD_2)$. Now the claim follows.
\end{proof}

\subsection{Proof of \theoremref{thrm:bound-population}}
\begin{proof}
    $\forall h \in \calH$,
    \begin{equation*}
        \begin{aligned}
            \varepsilon_{\target}(h) = \calL_{\target}(f_{\target}, h) & \le \calL_{\source}(f_{\target}, h) + \dd\left(\calD_{\source}, \calD_{\target}; \calH, \ell\right) \\
            & \le \dd\left(\calD_{\source}, \calD_{\target}; \calH, \ell\right) + \calL_{\source}(f_{\source}, f_{\target}) + \calL_{\source}(f_{\source}, h) \\
            & = \varepsilon_{\source}(h) + \dd\left(\calD_{\source}, \calD_{\target}; \calH, \ell\right) + \calL_{\source}(f_{\source}, f_{\target}).
        \end{aligned}
    \end{equation*}
    The first inequality is due to the definition of discrepancy distance and the assumption $f_{\target} \in \calH$. And the second inequality is according to the triangle inequality of $\ell$. Similarly, we have
    \begin{equation*}
        \begin{aligned}
            \varepsilon_{\target}(h) = \calL_{\target}(f_{\target}, h) & \le \calL_{\target}(f_{\source}, f_{\target}) + \calL_{\target}(f_{\source}, h) \\
            & \le \dd\left(\calD_{\source}, \calD_{\target}; \calH, \ell\right) + \calL_{\target}(f_{\source}, f_{\target}) + \calL_{\source}(f_{\source}, h) \\
            & = \varepsilon_{\source}(h) + \dd\left(\calD_{\source}, \calD_{\target}; \calH, \ell\right) + \calL_{\target}(f_{\source}, f_{\target}).
        \end{aligned}
    \end{equation*}
    Now the claim follows from the above two inequalities.
\end{proof}

\subsection{Proof of \theoremref{thrm:bound-population-minus}}
\begin{proof}
    $\forall h \in \calH$,
    $$
    \begin{aligned}
        \varepsilon_{\target}(h) = \calL_{\target}\left(f_{\target}, h\right) & \ge \calL_{\source}\left(f_\target, h\right) - \dd\left(\calD_{\source}, \calD_{\target}; \calH, \ell\right) \\
        & \ge \calL_\source\left(f_\source, f_\target\right) - \calL_\source\left(f_\source, h\right) - \dd\left(\calD_{\source}, \calD_{\target}; \calH, \ell\right) \\
        & = \calL_\source\left(f_\source, f_\target\right) - \dd\left(\calD_{\source}, \calD_{\target}; \calH, \ell\right) - \varepsilon_{\source}(h).
    \end{aligned}
    $$
    The first inequality is due to the definition of discrepancy distance and the assumption $f_{\target} \in \calH$. And the second inequality is according to the triangle inequality of $\ell$. Similarly, we have,
    $$
    \begin{aligned}
        \varepsilon_{\target}(h) = \calL_{\target}\left(f_{\target}, h\right) & \ge \calL_\target\left(f_\source, f_\target\right) - \calL_\target\left(f_\source, h\right) \\
        & \ge \calL_{\target}\left(f_\source, f_\target\right) - \dd\left(\calD_{\source}, \calD_{\target}; \calH, \ell\right) - \calL_\source\left(f_\source, h\right)\\
        & = \calL_{\target}\left(f_\source, f_\target\right) - \dd\left(\calD_{\source}, \calD_{\target}; \calH, \ell\right) - \varepsilon_{\source}(h).
    \end{aligned}
    $$
    Now the claim follows from the above two inequalities.
\end{proof}

\subsection{Proof of \theoremref{thrm:bound-empirical}}
\begin{proof}
    According to \theoremref{thrm:bound-population} and triangle inequality of $\dd(\cdot, \cdot; \calH, \ell)$ \citep{mansour2009domain},
    $$
    \begin{aligned}
        \begin{aligned}
            \varepsilon_{\target}(h) \le & \, \varepsilon_{\source}(h) +  \calM_{\covariate}\left(\calD_\source, \calD_\target; \calH, \ell\right) + \calM_{\concept}^{\min}\left(\calD_\source, \calD_\target, f_\source, f_\target; \ell\right) \\
            = & \, \varepsilon_{\source}(h) + \dd\left(\calD_{\source}, \calD_{\target}; \calH, \ell\right) + \min\left\{\calL_{\source}(f_{\source}, f_{\target}), \calL_{\target}(f_{\source}, f_{\target})\right\} \\
            \le & \, \varepsilon_{\source}(h) + \dd\left(\calD_{\source}, \hat{\calD}_{\source}; \calH, \ell\right) + \dd\left(\hat{\calD}_{\source}, \hat{\calD}_{\target}; \calH, \ell\right) + \dd\left(\hat{\calD}_{\target}, \calD_{\target}; \calH, \ell\right) \\
            & + \min\left\{\calL_{\source}(f_{\source}, f_{\target}), \calL_{\target}(f_{\source}, f_{\target})\right\}.
        \end{aligned}
    \end{aligned}
    $$
    According to \lemmaref{lemma:rademacher}, with probability at least $1 - \delta / 3$, $\forall h \in \calH$,
    $$
    \begin{aligned}
    \varepsilon_{\source}(h) = \calL_{\calD_{\source}}(h) & \le \hat{\calL}_{\source}(h) + \hat{\rad}_{S_\source}(\ell \circ \calH) + 3M\sqrt{\frac{\log(6 / \delta)}{2n_{\source}}} \\
    & = \hat{\varepsilon}_{\source}(h) + \hat{\rad}_{S_\source}(\ell \circ \calH) + 3M\sqrt{\frac{\log(6 / \delta)}{2n_{\source}}}.
    \end{aligned}
    $$
    In addition, according to \lemmaref{lemma:generalization-dd}, with probability at least $1 - \delta / 3$,
    $$
    \dd\left(\calD_{\source}, \hat{\calD}_{\source}; \calH, \ell\right) \le \hat{\rad}_{S_\source}(\calL_{\calH}) + 3M\sqrt{\frac{\log (6 / \delta)}{2n_{\source}}}.
    $$
    And with probability at least $1 - \delta / 3$,
    $$
    \dd\left(\calD_{\target}, \hat{\calD}_{\target}; \calH, \ell\right) \le \hat{\rad}_{S_\target}(\calL_{\calH}) + 3M\sqrt{\frac{\log (6 / \delta)}{2n_{\target}}}.
    $$
    Now the claim follows from the three inequalities above.
\end{proof}

\subsection{Proof of \theoremref{thrm:bound-empirical-minus}}
\begin{proof}
    According to \theoremref{thrm:bound-population-minus} and triangle inequality of $\dd(\cdot, \cdot; \calH, \ell)$ \citep{mansour2009domain},
    $$
    \begin{aligned}
        \varepsilon_{\target}(h) \ge & \, \calM_{\concept}^{\max}\left(\calD_\source, \calD_\target, f_\source, f_\target; \ell\right) - \calM_{\covariate}\left(\calD_\source, \calD_\target; \calH, \ell\right) - \varepsilon_{\source}(h) \\
        = & \, \max\left\{\calL_{\source}(f_{\source}, f_{\target}), \calL_{\target}(f_{\source}, f_{\target})\right\} - \dd\left(\calD_{\source}, \calD_{\target}; \calH, \ell\right) - \varepsilon_{\source}(h) \\
        \ge & \, \max\left\{\calL_{\source}(f_{\source}, f_{\target}), \calL_{\target}(f_{\source}, f_{\target})\right\} - \varepsilon_{\source}(h) \\
        & - \left(\dd\left(\calD_{\source}, \hat{\calD}_{\source}; \calH, \ell\right) + \dd\left(\hat{\calD}_{\source}, \hat{\calD}_{\target}; \calH, \ell\right) + \dd\left(\hat{\calD}_{\target}, \calD_{\target}; \calH, \ell\right)\right).
    \end{aligned}
    $$
    Similar to the proof of \theoremref{thrm:bound-empirical}, the claim follows from the forementioned three inequalities.
\end{proof}

\section{More Experiments and Discussions}
We present more experimental results and discussion about other backbones, pretraining methods and other split of NICO$^{++}$.

\subsection{Benchmark with ResNet-18 as Backbone}
As a large scale dataset, NICO$^{++}$ is diverse and rich enough to support training of ResNet-50 and ResNet-18. In the main paper we present Benchmark of classic DG and flexible DG with ResNet-50 as the backbone for current DG algorithms. In this section we benchmark current DG algorithms with ResNet-18 as the backbone. We keep the experimental settings and data split the same as those in Section 5.2 and 5.3 in the main paper and results of classic DG setting are in \tableref{tab:dg-appendix} and results of flexible DG setting are in \tableref{tab:ood-app}. 

Please note we adopt two methods to calculate the oracle results for with and without domain labels. Specifically, in the first approach we randomly split all data in target domains into training, validation and test sets with the ratio of 7:1:2 and train the model with ERM on the training subset, so that the model is trained with a mixture of target domains. In the second approach, we randomly split each target domain into training, validation and test sets with the ratio of 7:1:2, and train a model for each of target domains, so that both the training and test data are from a single domain in each training. We report the results of the first approach which is lower than the second approach in Table 2 and Table 3 in main paper donated as \textit{oracle}. We donate the results of the first approach as \textit{oracle} and the second as \textit{oracle*} here in \tableref{tab:dg-appendix}. 

SOTA methods including EoA, CORAL and StableNet still show outstanding performance with ResNet-18 as the backbone, which is consistent with results in Section 5.2 in the main paper, indicating the stability and consistency when benchmarking with NICO$^{++}$ across different backbones.

\begin{table*}[t]
    \centering
    \caption{Results of the DG setting on NICO$^{++}$. We report the accuracy on each target domain, overall accuracy, mean accuracy, and variance of accuracies across all target domains. We reimplement state-of-the-art unsupervised methods on NICO$^{++}$ with ResNet-18 as the backbone network for all the methods unless otherwise specified. Oracle donates the ResNet-18 trained with data sampled from the target distribution (yet none of test images is seen in the training). Ova. and Avg. indicate the overall accuracy of all the test data and the arithmetic mean of the accuracy of 3 domains, respectively. Note that they are different because the capacities of different domains are not equal. The reported results are average over three repetitions of each run. The best results of all methods are highlighted with the bold font.}
    \resizebox{1\textwidth}{!}{
    \begin{tabular}{c|cccc|ccc|cccc|ccc}
        \toprule
        \multirow{2}{*}{Method} & \multicolumn{7}{c|}{Training domains: G, Wa, R, A, I, Di} & \multicolumn{7}{c}{Training domains: S, G, Wa, R, I, O} \\
        \cmidrule{2-15}
        & S & Wi & O & Da & Ova. & Avg. & Std & A & Wi & Da & Di & Ova. & Avg. & Std \\
        \midrule
        Deepall   & 72.27 & 71.64  & 63.89   & 65.97 & 68.38   & 68.44   & 3.60 & 73.86  & 71.38  & 69.99 & 68.00 & 71.02   & 70.81   & 2.14 \\
        AdaClust  & 65.40 & 65.90  & 58.16   & 59.76 & 62.32   & 62.30   & 3.40 & 67.36  & 64.62  & 63.00 & 60.45 & 64.11   & 63.86   & 2.51 \\
        SagNet    & 71.76 & 70.90  & 63.54   & 64.88 & 67.72   & 67.77   & 3.61 & 74.04  & 71.08  & 70.05 & 67.96 & 71.00   & 70.78   & 2.19 \\
        EoA       & 74.12 & \textbf{73.78}  & 65.65   & 69.11 & 70.58   & 70.67   & 3.51 & 75.52  & 73.30  & 71.39 & 70.59 & 72.83   & 72.70   & 1.90 \\
        Mixstyle  & 72.25 & 70.73  & 63.55   & 65.63 & 67.92   & 68.04   & 3.57 & 73.28  & 70.53  & 66.82 & 67.52 & 70.33   & 69.54   & 2.57 \\
        MLDG      & 73.29 & 72.21  & 64.90   & 66.38 & 69.12   & 69.19   & 3.61 & 74.64  & 71.61  & 70.96 & 68.43 & 71.66   & 71.41   & 2.21 \\
        MMD       & 72.32 & 71.55  & 64.07   & 66.09 & 68.44   & 68.51   & 3.51 & 73.59  & 70.79  & 70.03 & 68.32 & 70.87   & 70.68   & 1.90 \\
        CORAL     & \textbf{74.77} & 73.50  & 66.43   & 68.97 & 70.80   & 70.92   & 3.37 & \textbf{75.84}  & \textbf{73.37}  & \textbf{72.12} & 71.04 & \textbf{73.23} & \textbf{73.09}   & 1.79 \\
        StableNet &   74.02    &    73.53    &  68.11       &  68.25     &    71.07     &  70.98      &    2.80  &   75.37     &  72.02      &     70.88  &    71.40   &    72.24     &     72.42    &   1.75   \\
        FACT      & 73.49 & 73.08  & \textbf{68.69}   & 69.62 & \textbf{71.19}   & 71.22   & \textbf{2.10} & 75.13  & 72.27  & 71.07 & 71.28 & 72.49   & 72.44   & \textbf{1.62} \\
        JiGen     & 74.10 & 72.88  & 68.41   & \textbf{69.75} & \textbf{71.19}   & \textbf{71.29}   & 2.30 & 75.04  & 72.59  & 70.74 & \textbf{71.42} & 72.47   & 72.45   & 1.64 \\
        GroupDRO  & 72.26 & 71.25  & 63.49   & 65.70 & 68.08   & 68.18   & 3.68 & 73.95  & 70.97  & 69.92 & 67.95 & 70.91   & 70.70   & 2.17 \\
        IRM       & 68.46 & 69.26  & 59.45   & 64.61 & 65.38   & 65.45   & 3.88 & 72.51  & 70.84  & 67.43 & 67.99 & 69.74   & 69.69   & 2.08 \\ \midrule
        Oracle & 81.53 & 82.21  & 78.34   & 78.57 & 80.22   & 80.16   & 1.73 & 82.23  & 82.83  & 77.19 & 80.51 & 80.54   & 80.69   & 2.19 \\
        Oracle*    & 85.69 & 84.26  & 82.22   & 82.92 & 83.72   & 83.77  & 1.33 & 85.51  & 84.26  & 82.92 & 82.85 & 83.93   & 83.88   & 1.09 \\ \bottomrule
    \end{tabular}}
    \tablelabel{tab:dg-appendix}
\end{table*}

\begin{table}[t]
    \centering
    \caption{Results of the flexible DG setting on NICO$^{++}$ with ResNet-18 as backbone.}
    \resizebox{\linewidth}{!}{
    \begin{tabular}{c|ccccccccccc|c}
        \toprule
        Method & Deepall & SWAD & MMLD & RSC & AdaClust & SagNet & EoA & MixStyle & StableNet & FACT & JiGen & Oracle \\
        \midrule
        Rand.  & 64.76   &  67.14    &   66.09   &  65.97   & 63.29    & 64.51  & 67.13 & 64.59    &     67.29      & \textbf{68.42} & 67.44 & 76.01  \\
        Comp.  & 68.93   &   70.25   &  68.20    &   68.22  & 66.33    & 68.43  & 70.85 & 67.86    &        70.72   & \textbf{71.70} & 70.64 & 78.63  \\
        Avg.   & 66.84   &   68.70   &  67.15    &  67.10   & 64.81    & 66.47  & 68.99 & 66.23    &       69.00    & \textbf{70.06} & 69.04 & 77.32  \\ \bottomrule
    \end{tabular}
    }
    \tablelabel{tab:ood-app}
\end{table}

\subsection{Pretraining Methods}
\sectionlabel{sec:pretrain}
Though the pretraining on ImageNet \citep{deng2009imagenet} is widely adopted in current visual recognition algorithms as the initialization of the model, the mapping from visual features to category labels can be biased and misleading given that ImageNet can be considered as a set of data sampled from latent domains \citep{shen2021towards,he2021towards} which can be different from those in a given DG benchmark. 
For example, the images in ImageNet are similar to the ones in domain \textit{photo} in PACS and \textit{real} in DomainNet while contrasting with other domains, so that ImageNet can be considered as an extension of specific domains, causing unbalance and bias in domains.
Moreover, if we consider the background of a image is its domain, then the diversity of background in ImageNet can leak knowledge about target domains which are supposed to be unknown in the training phase. Thus this is a critical problem in DG yet remains undiscussed.

We benchmark current DG methods with random initialization instead of pretrained on ImageNet. We adopt randomly initialized ResNet-50 as the backbone and keep the experimental settings and data split the same as those in Section 5.2 and 5.3 in the main paper.
The results are shown in \tableref{tab:dg_pretrain-app}. 
Without pretraining, both ERM and most current DG methods still show valid results. We fail to achieve valid results with IRM and MLDG, which may be caused by the requirement of careful tunning and subtle choice of hyperparameters.

\begin{table*}[t]
    \centering
    \caption{Results of the DG setting on NICO$^{++}$ with randomly initialized ResNet-50 as the backbone.}
    \resizebox{1\textwidth}{!}{
    \begin{tabular}{c|cccc|ccc|cccc|ccc}
        \toprule
        \multirow{2}{*}{Method} & \multicolumn{7}{c|}{Training domains: G, Wa, R, A, I, Di} & \multicolumn{7}{c}{Training domains: S, G, Wa, R, I, O} \\
        \cmidrule{2-15}
        & S & Wi & O & Da & Ova. & Avg. & Std & A & Wi & Da & Di & Ova. & Avg. & Std \\
        \midrule
        Deepall & 57.25 & 57.88 & 50.54 & 50.39 & 54.01 & 54.02 & 4.69 & 58.16 & 50.45 & 60.14 & 51.15 & 55.57 & 54.98 & 4.24  \\ 
        SagNet    & 58.85  & 58.46  & 55.38 & 50.03 & 55.85   & 55.68   & 3.53 & 59.23 & \textbf{55.30}  & 59.28   & 50.10 & 56.79   & 55.98   & 3.76 \\
        EoA       & 58.03   & 57.39   & 54.15  & 50.22 & 54.82    & 54.95    & 3.10 & 58.82  & 54.27   & 58.20    & 51.55  & 56.19    & 55.71    & 2.97 \\
        Mixstyle  & 56.40  & 56.34  & 54.03 & 49.46 & 54.21   & 54.06   & 2.82 & \textbf{60.29} & 54.35  & 59.07   & 50.34 & 56.65   & 56.01   & 3.96 \\
        MMD       & 55.22  & 54.76  & 52.47 & 46.69 & 52.45   & 52.29   & 3.39 & 58.15 & 51.76  & 57.93   & 46.12 & 54.34   & 53.49   & 4.97 \\
        CORAL     & 58.09  & 56.89  & 54.52 & 47.88 & 54.50   & 54.35   & 3.95 & 58.56 & 54.51 & 58.89   & 47.98 & 55.76   & 54.99   & 4.40 \\
        StableNet &    \textbf{59.02}    &   \textbf{59.58}     &    54.49   &    \textbf{52.15}   &     \textbf{56.30}    &    \textbf{56.31}     &  3.11    &   59.96    &    53.25    &    \textbf{61.14}     &    50.07   &    \textbf{56.87}     &   \textbf{56.11}     &  4.60    \\
        JiGen     & 57.28  & 55.68  & \textbf{55.78} & 51.32 & 55.06   & 55.02   & \textbf{2.23} & 58.17 & 54.01  & 56.28   & \textbf{51.74} & 55.40   & 55.05   & \textbf{2.41} \\
        GroupDRO  & 57.88  & 56.53  & 55.76 & 48.90 & 54.91   & 54.77   & 3.47 & 58.29 & 53.00  & 59.11   & 47.84 & 55.35   & 54.56   & 4.53 \\
         \bottomrule

    \end{tabular}}
    \tablelabel{tab:dg_pretrain-app}
\end{table*}

\begin{table}[t]
    \centering
    \caption{Results of the DG setting on NICO$^{++}$ with randomly initialized ResNet-50 as the backbone.}
    \resizebox{0.8\linewidth}{!}{
    \begin{tabular}{@{}c|ccccccccc@{}}
    \toprule
    Method & Deepall & SWAD & MMLD & RSC & SagNet & EoA  & MixStyle & StableNet & JiGen \\ \midrule
    Rand.  & 51.13   &   52.05   &  49.85    &  51.98   & 52.55  & 51.52 & 50.29    &      \textbf{52.95}     & 51.80 \\
    Comp.  & 53.39   &   \textbf{54.43}    & 53.27     &  53.11    & 53.71  & 53.79 & 53.92    &     53.28       & 54.21 \\
    Avg.   & 52.26   &  \textbf{53.24}    &    51.56  &  52.55   & 53.13  & 52.66 & 52.11    &     53.12     & 53.01 \\ \bottomrule
    \end{tabular}}
    
    \tablelabel{tab:ood_pretrain}
\end{table}

\subsection{Other Splits of Domains}
Given that NICO$^{++}$ contains 10 common domains and 10 unique domains, extensive experimental settings with controllable degree and type of contribution shifts can be constructed with various selection of domains for training and test data.
In the main paper we select \textit{grass, water, rock, autumn, indoor} and \textit{dim} as source domains and \textit{sand, winter, outdoor, dark} as target domains in the first split in Section 5.2 while \textit{autumn, winter, dark} and \textit{dim} as target domains and others as source domains in the second split. Here we benchmark DG methods with other split of training and testing domains. We randomly select \textit{rock, indoor, outdoor} and \textit{dim} for testing and others for training. The results are in \tableref{tab:other_split}. The consistency of outstanding performance of some SOTA methods including EoA, CORAL and StableNet across different splits indicates that the concept shifts between domains are comparable and small enough, so that common knowledge are strong and rich enough for models to learn. Please note the gap between \textit{oracle*} and \textit{oracle} is considerable and the improvement space on NICO$^{++}$ for DG methods is significant.

\begin{table*}[t]
    \centering
    \caption{Results of the DG setting on other split of NICO$^{++}$ with ImageNet pretrained ResNet-50 as the backbone. The training domains are \textit{grass, water, rock, autumn, indoor} and \textit{dim} while the others are test domains.}
    \resizebox{0.8\linewidth}{!}{
    \begin{tabular}{>{\centering}p{0.2\linewidth}|>{\centering}p{0.08\linewidth}>{\centering}p{0.08\linewidth}>{\centering}p{0.08\linewidth}>{\centering}p{0.08\linewidth}|>{\centering}p{0.08\linewidth}>{\centering}p{0.08\linewidth}c}
        \toprule
        \multirow{2}{*}{Method} & \multicolumn{7}{c}{Training domains: S, Wi, Da, G, Wa, A} \\
        \cmidrule{2-8}
        & R & I & O & Di & Ova. & Avg. & Std\\
        \midrule
        Deepall   & 79.87  & 58.18  & 77.39 & 74.91 & 72.79   & 72.59   & 8.50 \\
       
        AdaClust  & 78.51  & 55.72  & 75.34 & 72.72 & 70.76   & 70.57   & 8.82 \\
        SagNet    & 79.45  & 56.44  & 76.69 & 75.20 & 72.14   & 71.94   & 9.08 \\
        EoA       & 81.30  & \textbf{60.69}  & \textbf{78.75} & 76.06 & \textbf{74.39}   & \textbf{74.20}   & 8.02 \\
        Mixstyle  & 79.42  & 57.34  & 76.64 & 75.74 & 72.46   & 72.29   & 8.73 \\
        MLDG      & 80.13  & 59.03  & 77.49 & 75.23 & 73.15   & 72.97   & 8.23 \\
        MMD       & 80.60  & 59.15  & 77.96 & 75.73 & 73.55   & 73.36   & 8.38 \\
        CORAL     & \textbf{81.32}  & 59.52  & 78.44 & 76.64 & 74.15   & 73.98   & 8.51 \\
        StableNet &    80.98    &   59.88     &   78.65    &   76.11    &    74.11     &   73.91      &  8.28    \\
        FACT      & 79.89  & 57.53  & 77.27 & \textbf{77.63} & 73.25   & 73.08   & 9.03 \\
        JiGen     & 80.45  & 56.99  & 77.29 & 77.56 & 73.22   & 73.07   & 9.37 \\
        GroupDRO  & 80.06  & 58.44  & 77.62 & 75.21 & 73.04   & 72.83   & 8.49 \\
        IRM       & 70.19  & 48.96  & 66.16 & 61.76 & 61.90   & 61.77   & \textbf{7.97} \\
        \midrule
        Oracle    & 83.69  & 79.14  & 83.58 & 84.27 & 82.72   & 82.67   & 2.05 \\
        Oracle*    & 89.95  & 84.31  & 90.25 & 89.33 & 88.57   & 88.46  & 2.42 \\
        \bottomrule
    \end{tabular}}
    \tablelabel{tab:other_split}
\end{table*}

\subsection{Implementation Details}
\paragraph{Data generation.} The MNIST-M are generated by blending digit figures from the original MNIST dataset over patches extracted from images in BSDS500 dataset. The backgrounds are cropped from 200 images, resulting in 200 domains. The backgrounds from the same domain may be different given they are randomly cropped from the same image.

\paragraph{Datasets evaluation.} For experiments of datasets evaluation in Section 4.3 in the main paper, we adopt ResNet-50 \citep{he2016deep} as the backbone for NICO$^{++}$, PACS, DomainNet, VLCS, and Office-Home and shallower CNNs for MNIST-M as its image size is small. We show the structure of the used shallow CNNs in \tableref{tab:shallow}. We set the learning rate to 0.1 and batch to 64 for 20 epochs of training.

\paragraph{DG benchmarks.}\footnote{\scriptsize Both NICO$^{++}$ and the code for benchmarking can be found at
\url{https://github.com/xxgege/NICO-plus}.}
For experiments of benchmarking DG algorithms, we adopt weights pretrained on ImageNet as the initialization in Section 5.2, 5.3 and 5.4 in the main paper. The batch size is 192, the training epoch number is 60, learning rate is 2e-3 and decays to 2e-4 at epoch 30, and weight decay is 1e-3. For experiments without pretrained initialization in \sectionref{sec:pretrain}, the batch size is 192, the training epoch number is 90, learning rate is 2e-2 with a cosine decay process, and weight decay is 1e-4.

\section{More Statics and Example Images of NICO$^{++}$}
We show the detailed statistics of common and unique domains of the NICO$^{++}$ dataset in \tableref{tab:common_domain} and \tableref{tab:unique_domain}, respectively. We present all the names of unique domains and image numbers for each category.

We show example images of the common and unique domains in NICO$^{++}$ in \figureref{fig:common} and \figureref{fig:unique}, respectively. 

\clearpage

\begin{scriptsize}
\centering
\begin{longtable}{c|cccccccccc|c}
    \caption{Detailed statistics of common domains in the NICO$^{++}$ dataset.}
    \tablelabel{tab:common_domain} \\
    \toprule & \multicolumn{10}{|c|}{\textbf{Common Domains}}\\
    \midrule \textbf{Category} &water & grass & sand & rock & autumn & winter & indoor & outdoor & dim & dark &\textbf{Total} \\
     \midrule car & 306 & 321 & 244 & 285 & 206 & 348 & 386 & 402 & 300 & 386 & 3184 \\
     \midrule flower & 358 & 419 & 222 & 322 & 128 & 218 & 229 & 341 & 221 & 319 & 2777 \\
     \midrule penguin & 396 & 355 & 258 & 233 & 50 & 364 & 50 & 174 & 276 & 50 & 2206 \\
     \midrule camel & 328 & 263 & 330 & 83 & 50 & 296 & 80 & 220 & 214 & 98 & 1962 \\
     \midrule chair & 503 & 213 & 216 & 81 & 234 & 236 & 332 & 276 & 145 & 111 & 2347 \\
     \midrule monitor & 50 & 62 & 50 & 50 & 50 & 50 & 313 & 67 & 50 & 50 & 792 \\
     \midrule truck & 442 & 359 & 213 & 232 & 174 & 218 & 204 & 246 & 331 & 213 & 2632 \\
     \midrule tiger & 374 & 297 & 50 & 201 & 126 & 328 & 218 & 78 & 73 & 199 & 1944 \\
     \midrule wheat & 106 & 290 & 50 & 50 & 137 & 133 & 50 & 139 & 199 & 115 & 1269 \\
     \midrule sword & 71 & 173 & 66 & 193 & 50 & 57 & 178 & 87 & 89 & 50 & 1014 \\
     \midrule seal & 414 & 290 & 284 & 272 & 50 & 355 & 50 & 269 & 115 & 50 & 2149 \\
     \midrule wolf & 277 & 239 & 120 & 265 & 235 & 281 & 107 & 50 & 179 & 137 & 1890 \\
     \midrule lion & 253 & 460 & 270 & 256 & 125 & 246 & 236 & 50 & 294 & 278 & 2468 \\
     \midrule fish & 248 & 186 & 94 & 95 & 50 & 50 & 311 & 50 & 82 & 100 & 1266 \\
     \midrule dolphin & 340 & 88 & 118 & 50 & 50 & 50 & 114 & 310 & 176 & 54 & 1350 \\
     \midrule lifeboat & 543 & 125 & 189 & 123 & 50 & 118 & 151 & 375 & 94 & 100 & 1868 \\
     \midrule tank & 162 & 252 & 202 & 50 & 50 & 247 & 258 & 234 & 65 & 96 & 1616 \\
     \midrule corn & 155 & 195 & 68 & 50 & 186 & 78 & 150 & 186 & 151 & 152 & 1371 \\
     \midrule fishing rod & 492 & 223 & 313 & 249 & 190 & 317 & 195 & 379 & 265 & 69 & 2692 \\
     \midrule owl & 230 & 378 & 167 & 123 & 193 & 328 & 166 & 197 & 290 & 251 & 2323 \\
     \midrule sunflower & 198 & 327 & 124 & 97 & 54 & 165 & 63 & 209 & 289 & 216 & 1742 \\
     \midrule cow & 387 & 861 & 323 & 150 & 233 & 445 & 296 & 263 & 268 & 128 & 3354 \\
     \midrule bird & 606 & 595 & 229 & 301 & 180 & 423 & 176 & 203 & 414 & 149 & 3276 \\
     \midrule clock & 213 & 283 & 182 & 84 & 252 & 259 & 239 & 267 & 94 & 171 & 2044 \\
     \midrule shrimp & 260 & 190 & 117 & 50 & 50 & 50 & 86 & 50 & 50 & 56 & 959 \\
     \midrule goose & 278 & 391 & 106 & 57 & 146 & 154 & 87 & 349 & 193 & 50 & 1811 \\
     \midrule airplane & 256 & 276 & 281 & 268 & 71 & 295 & 243 & 345 & 229 & 221 & 2485 \\
     \midrule shark & 289 & 123 & 209 & 50 & 50 & 50 & 52 & 257 & 255 & 162 & 1497 \\
     \midrule rabbit & 160 & 457 & 232 & 122 & 126 & 342 & 309 & 167 & 88 & 67 & 2070 \\
     \midrule snake & 252 & 364 & 347 & 206 & 150 & 74 & 197 & 187 & 50 & 142 & 1969 \\
     \midrule hot air balloon & 460 & 270 & 319 & 254 & 147 & 328 & 50 & 367 & 227 & 291 & 2713 \\
     \midrule lizard & 369 & 374 & 312 & 344 & 130 & 57 & 161 & 346 & 50 & 106 & 2249 \\
     \midrule hat & 280 & 285 & 295 & 73 & 210 & 142 & 376 & 404 & 147 & 92 & 2304 \\
     \midrule spider & 246 & 268 & 339 & 98 & 50 & 88 & 179 & 248 & 194 & 212 & 1922 \\
     \midrule motorcycle & 390 & 350 & 265 & 266 & 258 & 220 & 285 & 347 & 331 & 239 & 2951 \\
     \midrule tortoise & 292 & 357 & 300 & 199 & 68 & 50 & 134 & 291 & 64 & 50 & 1805 \\
     \midrule dog & 886 & 488 & 410 & 240 & 311 & 831 & 437 & 456 & 322 & 239 & 4620 \\
     \midrule crocodile & 343 & 255 & 272 & 151 & 50 & 50 & 138 & 327 & 77 & 157 & 1820 \\
     \midrule elephant & 402 & 455 & 326 & 85 & 50 & 169 & 96 & 286 & 338 & 168 & 2375 \\
     \midrule chicken & 210 & 268 & 138 & 50 & 80 & 291 & 211 & 272 & 51 & 50 & 1621 \\
     \midrule bee & 155 & 226 & 104 & 50 & 50 & 59 & 50 & 146 & 50 & 50 & 940 \\
     \midrule gun & 290 & 283 & 51 & 71 & 73 & 130 & 346 & 224 & 91 & 160 & 1719 \\
     \midrule fox & 186 & 401 & 236 & 152 & 217 & 271 & 172 & 161 & 133 & 193 & 2122 \\
     \midrule phone & 417 & 219 & 340 & 130 & 100 & 156 & 284 & 234 & 106 & 311 & 2297 \\
     \midrule bus & 348 & 332 & 195 & 187 & 162 & 262 & 280 & 367 & 202 & 220 & 2555 \\
     \midrule cat & 353 & 455 & 238 & 187 & 224 & 699 & 518 & 249 & 241 & 228 & 3392 \\
     \midrule sailboat & 434 & 332 & 222 & 236 & 92 & 226 & 78 & 402 & 251 & 205 & 2478 \\
     \midrule giraffe & 368 & 444 & 247 & 149 & 89 & 117 & 135 & 214 & 277 & 86 & 2126 \\
     \midrule cactus & 298 & 319 & 299 & 205 & 50 & 202 & 306 & 203 & 310 & 211 & 2403 \\
     \midrule pumpkin & 212 & 236 & 129 & 75 & 289 & 64 & 137 & 240 & 167 & 89 & 1638 \\
     \midrule train & 271 & 346 & 212 & 219 & 243 & 279 & 115 & 202 & 238 & 263 & 2388 \\
     \midrule dragonfly & 226 & 447 & 138 & 188 & 94 & 50 & 50 & 250 & 291 & 80 & 1814 \\
     \midrule ship & 402 & 203 & 225 & 205 & 74 & 213 & 200 & 378 & 302 & 244 & 2446 \\
     \midrule helicopter & 249 & 308 & 338 & 225 & 73 & 241 & 287 & 436 & 314 & 233 & 2704 \\
     \midrule bicycle & 327 & 362 & 215 & 327 & 208 & 321 & 202 & 415 & 385 & 253 & 3015 \\
     \midrule racket & 135 & 241 & 113 & 50 & 50 & 53 & 162 & 207 & 76 & 64 & 1151 \\
     \midrule squirrel & 209 & 437 & 299 & 241 & 272 & 376 & 80 & 266 & 107 & 91 & 2378 \\
     \midrule bear & 550 & 665 & 154 & 193 & 145 & 624 & 239 & 131 & 164 & 132 & 2997 \\
     \midrule scooter & 132 & 240 & 103 & 110 & 179 & 130 & 119 & 222 & 71 & 99 & 1405 \\
     \midrule mailbox & 92 & 309 & 227 & 234 & 89 & 239 & 73 & 229 & 78 & 50 & 1620 \\
     \midrule horse & 305 & 438 & 386 & 174 & 239 & 319 & 293 & 375 & 318 & 162 & 3009 \\
     \midrule pineapple & 363 & 240 & 249 & 50 & 50 & 63 & 125 & 154 & 50 & 59 & 1403 \\
     \midrule banana & 116 & 367 & 50 & 50 & 50 & 50 & 184 & 130 & 50 & 50 & 1097 \\
     \midrule mushroom & 96 & 321 & 155 & 50 & 254 & 111 & 173 & 245 & 99 & 129 & 1633 \\
     \midrule cauliflower & 84 & 79 & 50 & 50 & 50 & 50 & 119 & 79 & 50 & 50 & 661 \\
     \midrule whale & 222 & 87 & 205 & 60 & 50 & 103 & 50 & 214 & 282 & 73 & 1346 \\
     \midrule frog & 296 & 351 & 233 & 258 & 208 & 99 & 50 & 106 & 54 & 248 & 1903 \\
     \midrule football & 140 & 235 & 306 & 50 & 60 & 133 & 101 & 278 & 163 & 50 & 1516 \\
     \midrule camera & 254 & 255 & 253 & 126 & 249 & 208 & 275 & 211 & 261 & 139 & 2231 \\
     \midrule ostrich & 252 & 286 & 310 & 113 & 50 & 163 & 118 & 336 & 153 & 50 & 1831 \\
     \midrule beetle & 170 & 295 & 258 & 214 & 114 & 53 & 50 & 138 & 65 & 109 & 1466 \\
     \midrule tent & 441 & 389 & 270 & 250 & 265 & 279 & 163 & 288 & 280 & 288 & 2913 \\
     \midrule kangaroo & 252 & 346 & 304 & 110 & 76 & 250 & 102 & 197 & 257 & 120 & 2014 \\
     \midrule monkey & 251 & 322 & 139 & 337 & 93 & 222 & 253 & 231 & 184 & 99 & 2131 \\
     \midrule crab & 178 & 287 & 242 & 184 & 50 & 50 & 144 & 128 & 117 & 124 & 1504 \\
     \midrule lemon & 235 & 312 & 54 & 50 & 60 & 50 & 94 & 131 & 50 & 50 & 1086 \\
     \midrule pepper & 142 & 134 & 50 & 50 & 128 & 50 & 50 & 123 & 50 & 50 & 827 \\
     \midrule sheep & 292 & 438 & 237 & 335 & 273 & 239 & 329 & 395 & 303 & 135 & 2976 \\
     \midrule butterfly & 111 & 388 & 159 & 255 & 132 & 76 & 58 & 248 & 182 & 82 & 1691 \\
     \midrule umbrella & 364 & 303 & 238 & 119 & 232 & 208 & 196 & 372 & 250 & 246 & 2528 \\
     \bottomrule 
\end{longtable}
\end{scriptsize}

\clearpage

\newcolumntype{C}{>{\centering\let\newline\\\arraybackslash\hspace{0pt}}m{0.055\linewidth}}
\newcolumntype{P}[1]{>{\centering\let\newline\\\arraybackslash\hspace{0pt}}m{#1}}

\captionsetup{font=normalsize}
\begin{scriptsize}
\begin{longtable}{P{0.1\linewidth}|CCCCCCCCCC|P{0.05\linewidth}}
    \caption{\normalsize Detailed statistics of unique domains in the NICO$^{++}$ dataset.}
    \tablelabel{tab:unique_domain} \\
    \toprule 
    \textbf{Category} & \multicolumn{10}{|c|}{\textbf{Unique Domains}} & \textbf{Total}\\
    \midrule 
     \multirow{2}{*}{car} & red & green & on track & across bridge & repairing & aside people & in gas station & without roof & on booth & aside traffic light &  \multirow{2}{*}{669} \\ \cmidrule{2-11}
     & 139 & 114 & 77 & 77 & 57 & 51 & 47 & 40 & 37 & 30 &    \\
    \midrule 
     \multirow{2}{*}{flower} & peony & in vase & bouquet & carnation & rose & in glass dome & chrysanthemum & holding & wreath & on ear &  \multirow{2}{*}{1073} \\ \cmidrule{2-11}
     & 140 & 133 & 132 & 125 & 122 & 122 & 115 & 89 & 65 & 30 &    \\
    \midrule 
     \multirow{2}{*}{penguin} & with hair & brown & lying & blue & in mud & watching egg & in cave & opening mouth & with shadow & with child &  \multirow{2}{*}{402} \\ \cmidrule{2-11}
     & 62 & 56 & 53 & 47 & 34 & 30 & 30 & 30 & 30 & 30 &    \\
    \midrule 
     \multirow{2}{*}{camel} & people riding & sitting & lying & carrying goods & white & with single hump & on leash & roaring & with triple humps & in cave &  \multirow{2}{*}{698} \\ \cmidrule{2-11}
     & 125 & 124 & 93 & 87 & 80 & 69 & 30 & 30 & 30 & 30 &    \\
    \midrule 
     \multirow{2}{*}{chair} & wooden & armchair & rocking chair & with cushion & circle & lying & people sitting on & green & in classroom & red & \multirow{2}{*}{959} \\ \cmidrule{2-11}
     & 137 & 132 & 124 & 117 & 107 & 98 & 94 & 90 & 30 & 30 &   \\
    \midrule 
     \multirow{2}{*}{monitor} & ultrawide & curved & beside keyboard & white & touching & beside laptop & micro & in box & on table & turned off &  \multirow{2}{*}{426} \\ \cmidrule{2-11}
     & 93 & 63 & 52 & 38 & 30 & 30 & 30 & 30 & 30 & 30 &   \\
    \midrule 
     \multirow{2}{*}{truck} & abandon & with crane & carrying contrainer & yellow & repairing & armed & in gas station & in race & out of  tunnel & without container &  \multirow{2}{*}{784} \\ \cmidrule{2-11}
     & 122 & 119 & 111 & 105 & 81 & 73 & 58 & 52 & 33 & 30 &   \\
    \midrule 
     \multirow{2}{*}{tiger} & lying & white & eating & roaring & passing the ring of fire & in cave & in mud & with shadow & with chain & in hospital &  \multirow{2}{*}{648} \\ \cmidrule{2-11}
     & 143 & 128 & 121 & 54 & 41 & 41 & 30 & 30 & 30 & 30 &    \\
    \midrule 
     \multirow{2}{*}{wheat} & ear of wheat & green & being harvested & wheat on hand & in jar & on table & hanging & in mouth & tied up by red ribbon & through  magnifier & \multirow{2}{*}{697} \\ \cmidrule{2-11}
     & 142 & 139 & 117 & 97 & 52 & 30 & 30 & 30 & 30 & 30 &    \\
    \midrule 
     \multirow{2}{*}{sword} & wooden & holding & dagger & on rack & in scabbard & fencing & golden & with shield & with tassel & in mud &  \multirow{2}{*}{684} \\ \cmidrule{2-11}
     & 123 & 105 & 104 & 100 & 81 & 41 & 40 & 30 & 30 & 30 &    \\
    \midrule 
     \multirow{2}{*}{seal} & spotted & in aquarium & white & belly up & standing & playing with ball & diving & sitting & grey & with baby &  \multirow{2}{*}{502} \\ \cmidrule{2-11}
     & 119 & 79 & 72 & 42 & 35 & 35 & 30 & 30 & 30 & 30 &    \\
    \midrule 
     \multirow{2}{*}{wolf} & white & running & cub wolf & in cave & roaring & in mud & stick outing  tongue & with shadow & belly up & under moon &  \multirow{2}{*}{559} \\ \cmidrule{2-11}
     & 127 & 124 & 98 & 30 & 30 & 30 & 30 & 30 & 30 & 30 &   \\
    \midrule 
     \multirow{2}{*}{lion} & cub lion & sleeping & running & eating & white & lioness & roaring & in mud & in cave & preying on hippo &  \multirow{2}{*}{800} \\ \cmidrule{2-11}
     & 132 & 131 & 127 & 124 & 85 & 61 & 50 & 30 & 30 & 30 &   \\
    \midrule 
     \multirow{2}{*}{fish} & black goldfish & opening mouth & in tank & glowing & red crucian & in net & on hand & in ice & with baby & eating &  \multirow{2}{*}{429} \\ \cmidrule{2-11}
     & 118 & 57 & 44 & 30 & 30 & 30 & 30 & 30 & 30 & 30 &    \\
    \midrule 
     \multirow{2}{*}{dolphin} & playing with ball & jumping & in aquarium & white & black & through ring & with baby & standing& diving & aside people & \multirow{2}{*}{617} \\ \cmidrule{2-11}
     & 134 & 134 & 114 & 53 & 32 & 30 & 30 & 30 & 30 & 30 &    \\
    \midrule 
     \multirow{2}{*}{lifeboat} & with people & hanging & enclosed & on wave & yellow & rubber & with paddle & white & across bridge & repairing &\multirow{2}{*}{662} \\ \cmidrule{2-11}
     & 120 & 107 & 102 & 85 & 83 & 43 & 32 & 30 & 30 & 30 &    \\
    \midrule 
     \multirow{2}{*}{tank} & with soldier & firing & with air defense & amphibious & carrying missile & in smoke & in swamp & with flag & green & turn over &  \multirow{2}{*}{577} \\ \cmidrule{2-11}
     & 106 & 101 & 93 & 84 & 37 & 34 & 32 & 30 & 30 & 30 &    \\
    \midrule 
     \multirow{2}{*}{corn} & holding & in basket & eating & red & eaten & with cob & on a stick & with leaf & colorful & roasted & \multirow{2}{*}{946} \\ \cmidrule{2-11}
     & 143 & 136 & 121 & 100 & 99 & 81 & 81 & 74 & 58 & 53 &    \\
    \midrule 
     \multirow{2}{*}{fishing rod} & on rack & on hand & wooden & blue & straight & in bucket & on railing & with winding wheel & curved & in bag & \multirow{2}{*}{618} \\ \cmidrule{2-11}
     & 108 & 89 & 75 & 74 & 59 & 58 & 53 & 42 & 30 & 30 &   \\
    \midrule 
     \multirow{2}{*}{owl} & sleeping & flying & white & lying & preying & in cave & on shoulder & under moon & running & on arm &  \multirow{2}{*}{555} \\ \cmidrule{2-11}
     & 123 & 117 & 94 & 36 & 35 & 30 & 30 & 30 & 30 & 30 &  \\
    \midrule 
     \multirow{2}{*}{sunflower} & with sunglass & under sun & red & wilted & potted & white & in glass dome & aside people & beside windmill & with cloud & \multirow{2}{*}{738} \\ \cmidrule{2-11}
     & 144 & 118 & 117 & 101 & 82 & 55 & 31 & 30 & 30 & 30 &    \\
    \midrule 
     \multirow{2}{*}{cow} & lying & baby cow & being milked & Indian cow & with curly hair & with long horn & spotted & aside people & jumping & on steroids & \multirow{2}{*}{813} \\ \cmidrule{2-11}
     & 137 & 125 & 117 & 117 & 77 & 60 & 57 & 48 & 45 & 30&   \\
    \midrule 
     \multirow{2}{*}{bird} & long beak & yellow & flying & on hand & green & opening mouth & eating & in nest & on shoulder & walking & \multirow{2}{*}{804} \\ \cmidrule{2-11}
     & 114 & 112 & 99 & 97 & 83 & 81 & 76 & 73 & 39 & 30 &    \\
    \midrule 
     \multirow{2}{*}{clock} & mechanical watch & pendulum clock & alarm & pocket watch & timer & on tower & on wall & electric & on table & on arm & \multirow{2}{*}{847} \\ \cmidrule{2-11}
     & 121 & 118 & 110 & 108 & 96 & 95 & 64 & 58 & 47 & 30 &   \\
    \midrule 
     \multirow{2}{*}{shrimp} & on hand & transparent & cooked & in net & dark Brown-shelled Shrimp & lobster & in ice & giving birth & glowing & eating & \multirow{2}{*}{334} \\ \cmidrule{2-11}
     & 64 & 30 & 30 & 30 & 30 & 30 & 30 & 30 & 30 & 30 &   \\
    \midrule 
     \multirow{2}{*}{goose} & flapping wings & in wetland & eating & hatching & in mud & on  roof & black & being caught & in egg & aside people &\multirow{2}{*}{441} \\ \cmidrule{2-11}
     & 114 & 57 & 44 & 39 & 37 & 30 & 30 & 30 & 30 & 30 &   \\
    \midrule 
     \multirow{2}{*}{airplane} & taking off & fighter & biplane & with plane ladder & civil & with rainbow & aside pilot & on ship & with cloud & with the sun & \multirow{2}{*}{671} \\ \cmidrule{2-11}
     & 130 & 124 & 95 & 90 & 60 & 52 & 30 & 30 & 30 & 30 &   \\
    \midrule 
     \multirow{2}{*}{shark} & great white shark & opening mouth & in aquarium & belly up & being preyed & hardback dwarf shark & preying & diving & wounded & beside cage & \multirow{2}{*}{492} \\ \cmidrule{2-11}
     & 117 & 85 & 76 & 34 & 30 & 30 & 30 & 30 & 30 & 30 &   \\
    \midrule 
     \multirow{2}{*}{rabbit} & red eye & eating carrot & black & jumping & angus rabbit & on hand & with clother & with ribbon & in cave & belly up & \multirow{2}{*}{668} \\ \cmidrule{2-11}
     & 137 & 128 & 124 & 95 & 62 & 32 & 30 & 30 & 30 & 30 &   \\
    \midrule 
     \multirow{2}{*}{snake} & eating & sticking out tongue & in hole & white & circling & in egg & attacking & on hand & cobra & on stick &\multirow{2}{*}{562} \\ \cmidrule{2-11}
     & 106 & 99 & 78 & 57 & 52 & 50 & 30 & 30 & 30 & 30 &   \\
    \midrule 
     \multirow{2}{*}{hot air balloon} & yellow & on fire & on ground & nearby tower & festival & black & pink & with rainbow & red & black & \multirow{2}{*}{367} \\ \cmidrule{2-11}
     & 100 & 74 & 37 & 32 & 32 & 32 & 30 & 30 & 30 & 30 &   \\
    \midrule 
     \multirow{2}{*}{lizard} & sticking out tongue & on hand & orange & eating worms & in cave & in mud & green & on stick & standing & preying & \multirow{2}{*}{481} \\ \cmidrule{2-11}
     & 127 & 126 & 120 & 48 & 30 & 30 & 30 & 30 & 30 & 30 &   \\
    \midrule 
     \multirow{2}{*}{hat} & straw hat & top hat & blue & with mask & woolen & hanging & helmet & woolly & besides sunglass & flat cap & \multirow{2}{*}{836} \\ \cmidrule{2-11}
     & 125 & 112 & 107 & 105 & 94 & 88 & 63 & 52 & 30 & 30 & 30   \\
    \midrule 
     \multirow{2}{*}{spider} & hairy & yellow & on hand & spining silk & specimen & white & in spider web & in hole & lying & crawl &  \multirow{2}{*}{711} \\ \cmidrule{2-11}
     & 116 & 109 & 99 & 81 & 78 & 74 & 52 & 42 & 30 & 30   \\
    \midrule 
     \multirow{2}{*}{motorcycle} & repairing & on track & red & in gas station & aside people & abandon & with container & with shade & opening headlight & aside traffic light &  \multirow{2}{*}{706} \\ \cmidrule{2-11}
     & 139 & 125 & 123 & 71 & 64 & 54 & 40 & 30 & 30 & 30 &    \\
    \midrule 
     \multirow{2}{*}{tortoise} & on hand & belly up & in cave & green & eating earthworms & in net & mouth opened & carrying baby & carrying box & with people &  \multirow{2}{*}{337} \\ \cmidrule{2-11}
     & 61 & 36 & 30 & 30 & 30 & 30 & 30 & 30 & 30 & 30 &   \\
    \midrule 
     \multirow{2}{*}{dog} & lying & pug dog & wearing clothes & running & with dog chain & teddy dog & eating & on stairs & in cave & stick outing  tongue &  \multirow{2}{*}{995} \\ \cmidrule{2-11}
     & 144 & 137 & 127 & 121 & 112 & 107 & 98 & 89 & 30 & 30 &  \\
    \midrule 
     \multirow{2}{*}{crocodile} & preying & tied mouth & forest & in cage & aside people & in cave & on tile & belly up & in egg & wounded  & \multirow{2}{*}{317} \\ \cmidrule{2-11}
     & 50 & 37 & 30 & 30 & 30 & 30 & 30 & 30 & 30 & 30 &  \\
    \midrule 
     \multirow{2}{*}{elephant} & spraying water & in mud & baby. elephant & standing & in circus & sleeping & head of elephant & aside people & white elephant & wearing clothes &  \multirow{2}{*}{764} \\ \cmidrule{2-11}
     & 132 & 109 & 109 & 96 & 95 & 73 & 56 & 34 & 30 & 30 &   \\
    \midrule 
     \multirow{2}{*}{chicken} & black & running & flying & hatching & crowing & laying eggs & on hand & being caught & eating & in mud &  \multirow{2}{*}{529} \\ \cmidrule{2-11}
     & 106 & 83 & 69 & 64 & 48 & 39 & 30 & 30 & 30 & 30 &    \\
    \midrule 
     \multirow{2}{*}{bee} & flying & in hive & in honey comb & green & in hole & on hand & in jar & attacking & lying & on net & \multirow{2}{*}{597} \\ \cmidrule{2-11}
     & 106 & 103 & 88 & 80 & 65 & 35 & 30 & 30 & 30 & 30 &  \\
    \midrule 
     \multirow{2}{*}{gun} & small pistol & long-barrelled & air rifle & in holster & on table & firing & with sighting mirror & with bullet belt & raise & on armrack &  \multirow{2}{*}{558} \\ \cmidrule{2-11}
     & 120 & 101 & 66 & 64 & 49 & 37 & 31 & 30 & 30 & 30 &   \\
    \midrule 
     \multirow{2}{*}{fox} & with big ear & baby & white & running & eating & sitting & sleeping & with people & in cave & roaring &  \multirow{2}{*}{785} \\ \cmidrule{2-11}
     & 134 & 122 & 111 & 105 & 88 & 81 & 54 & 30 & 30 & 30 &    \\
    \midrule 
     \multirow{2}{*}{phone} & in hand & calling & foldable & beside laptop & on tripod & keyboard & inside pocket & slide & beside pillow & on table &  \multirow{2}{*}{780} \\ \cmidrule{2-11}
     & 135 & 128 & 115 & 114 & 88 & 80 & 30 & 30 & 30 & 30 &   \\
    \midrule 
     \multirow{2}{*}{bus} & double-decker bus & articulated buses & school bus & across bridge & aside station & in gas station & trolley buses & aside traffic light & on zebra crossing & at toll station &  \multirow{2}{*}{543} \\ \cmidrule{2-11}
     & 124 & 108 & 53 & 49 & 47 & 41 & 31 & 30 & 30 & 30 &  \\
    \midrule 
     \multirow{2}{*}{cat} & walking & ragdoll cat & maine cat & eating & jumping & in bag & beside laptop & in cave & washing face & in mud &  \multirow{2}{*}{857} \\ \cmidrule{2-11}
     & 126 & 122 & 120 & 119 & 113 & 85 & 64 & 47 & 31 & 30 &    \\
    \midrule 
     \multirow{2}{*}{sailboat} & ketch & colorful sails & with awning & single sail & sloop & on wave & barque & aside people & across  bridge & racing &   \multirow{2}{*}{842} \\ \cmidrule{2-11}
     & 124 & 113 & 108 & 106 & 101 & 86 & 86 & 53 & 35 & 30 &  \\
    \midrule 
     \multirow{2}{*}{giraffe} & sitting & head of giraffe & running & being fed & white & sleeping & in cave &  tongue out & drinking & with baby & \multirow{2}{*}{723} \\ \cmidrule{2-11}
     & 132 & 132 & 124 & 82 & 78 & 55 & 30 & 30 & 30 & 30 &   \\
    \midrule 
     \multirow{2}{*}{cactus} & flowering & in flowerpot & columnar & with white hair & with red thorns & blue & flaky & cactus without thorns & spheroidal & touched by hand & \multirow{2}{*}{695} \\ \cmidrule{2-11}
     & 127 & 122 & 120 & 82 & 68 & 52 & 34 & 30 & 30 & 30 &   \\
    \midrule 
     \multirow{2}{*}{pumpkin} & green & top view & half & white & on hand & halloween & Spherical & holloween & with leaf & columnar & \multirow{2}{*}{555} \\ \cmidrule{2-11}
     & 106 & 97 & 84 & 59 & 47 & 40 & 32 & 30 & 30 & 30 &   \\
    \midrule 
     \multirow{2}{*}{train} & steam train & people getting on off & tram & maglev & on bridge & subway & green & head of train & at station & cross tunnel & \multirow{2}{*}{962} \\ \cmidrule{2-11}
     & 127 & 117 & 113 & 112 & 107 & 106 & 89 & 83 & 78 & 30 &   \\
    \midrule 
     \multirow{2}{*}{dragonfly} & blue & side view & on rope & flying & specimen & pink & on hand & be preying & white & on  bricks & \multirow{2}{*}{684} \\ \cmidrule{2-11}
     & 123 & 103 & 83 & 80 & 74 & 68 & 56 & 35 & 32 & 30 &  \\
    \midrule 
     \multirow{2}{*}{ship} & cruise & military & cargo ship & anchored & with flag & with steam & sinking & green & with spray & civil &  \multirow{2}{*}{682} \\ \cmidrule{2-11}
     & 123 & 116 & 106 & 72 & 71 & 46 & 46 & 42 & 30 & 30 &   \\
    \midrule 
     \multirow{2}{*}{helicopter} & combat helicopter & small chopper & landing & camouflage & aside pilot & smoky & transport & landed & clipart & diving & \multirow{2}{*}{397} \\ \cmidrule{2-11}
     & 121 & 88 & 74 & 48 & 36 & 30 & 30 & 30 & 30 & 30 &   \\
    \midrule 
     \multirow{2}{*}{bicycle} & repairing & yellow & tandem & with training wheel & in velodrome & green & electric & aside people & with container & aside traffic light & \multirow{2}{*}{898} \\ \cmidrule{2-11}
     & 142 & 136 & 125 & 120 & 111 & 92 & 60 & 52 & 30 & 30 &   \\
    \midrule 
     \multirow{2}{*}{racket} & with tennis ball & broken & on hand & wooden & blue & racket in front of face & white & hanging & with badminton & in bag & \multirow{2}{*}{798} \\ \cmidrule{2-11}
     & 132 & 129 & 124 & 105 & 95 & 55 & 48 & 45 & 35 & 30 &   \\
    \midrule 
     \multirow{2}{*}{squirrel} & eating & black & on hand & fat & lying & jumping & in hole & on table & hanging & carrying cone & \multirow{2}{*}{999} \\ \cmidrule{2-11}
     & 131 & 128 & 122 & 117 & 114 & 109 & 101 & 73 & 71 & 33 &   \\
    \midrule 
     \multirow{2}{*}{bear} & lying & in cage & brown & polar bear & black & wombat & roaring & sitting & panda & teddy bear & \multirow{2}{*}{1081} \\ \cmidrule{2-11}
     & 138 & 137 & 130 & 128 & 125 & 119 & 102 & 92 & 80 & 30 &   \\
    \midrule 
     \multirow{2}{*}{scooter} & with child & blue & white & pink & double wheel & triple wheel & folded & on zebra crossing & with basket & swings & \multirow{2}{*}{529} \\ \cmidrule{2-11}
     & 100 & 84 & 71 & 69 & 43 & 41 & 31 & 30 & 30 & 30 &   \\
    \midrule 
     \multirow{2}{*}{mailbox} & red & green & wooden & open & with flag & square & with lamp & closed & columnar & aside people & \multirow{2}{*}{689} \\ \cmidrule{2-11}
     & 137 & 124 & 110 & 99 & 60 & 39 & 30 & 30 & 30 & 30 &   \\
    \midrule 
     \multirow{2}{*}{horse} & lying & running & carriage & racing & with saddle & opening mouth & pony & aside people & across hurdle & kissed by  people &  \multirow{2}{*}{787} \\ \cmidrule{2-11}
     & 130 & 127 & 118 & 78 & 77 & 66 & 60 & 58 & 43 & 30 &   \\
    \midrule 
     \multirow{2}{*}{pineapple} & peeled pineapple & with sunglasses & rotten & people eating pineapple & grilled & being cutted & on stick & in baskets & green & in bag & \multirow{2}{*}{561} \\ \cmidrule{2-11}
     & 115 & 113 & 54 & 54 & 45 & 45 & 44 & 31 & 30 & 30    \\
    \midrule 
     \multirow{2}{*}{banana} & unripe banana & peeled banana & in hand & people eating banana & fried & on stick & with fork & broken & in baskets & in bag & \multirow{2}{*}{669} \\ \cmidrule{2-11}
     & 136 & 135 & 100 & 87 & 52 & 39 & 30 & 30 & 30 & 30 &  \\
    \midrule 
     \multirow{2}{*}{mushroom} & red & purple & flammulina velutipes & lentinus edodes & russula lactea? & dehydrated & tricholoma & in basket & pleurotus eryngii & green & \multirow{2}{*}{887} \\ \cmidrule{2-11}
     & 142 & 131 & 111 & 94 & 93 & 84 & 75 & 62 & 60 & 35 &   \\
    \midrule 
     \multirow{2}{*}{cauliflower} & romanesco broccoli & purple & sprouting broccoli & with leaf & in basket & cooked & on plate & orange & on hand & in pot & \multirow{2}{*}{717} \\ \cmidrule{2-11}
     & 139 & 121 & 82 & 80 & 79 & 67 & 49 & 38 & 32 & 30 &   \\
    \midrule 
     \multirow{2}{*}{whale} & white & opening mouth & blue & spraying water & with baby & jumping & be preyed & diving & wounded & belly up & \multirow{2}{*}{470} \\ \cmidrule{2-11}
     & 87 & 77 & 73 & 53 & 30 & 30 & 30 & 30 & 30 & 30 &   \\
    \midrule 
     \multirow{2}{*}{frog} & on lotus leaf & in mud & preying & breathing & jumping & chocolate frog & red eyes & on hand & in cage & black eyes &  \multirow{2}{*}{471} \\ \cmidrule{2-11}
     & 113 & 95 & 39 & 38 & 36 & 30 & 30 & 30 & 30 & 30 &    \\
    \midrule 
     \multirow{2}{*}{football} & kicking & heading & in mud & deflated & goal & on hand & in bag & gold & colorful & on head &  \multirow{2}{*}{421} \\ \cmidrule{2-11}
     & 89 & 58 & 55 & 39 & 30 & 30 & 30 & 30 & 30 & 30 &   \\
    \midrule 
     \multirow{2}{*}{camera} & on hand & on tripod & polaroid & on ceiling & long lens camera & hanging & green & in bag & dual lens camera & flashing & \multirow{2}{*}{803} \\ \cmidrule{2-11}
     & 120 & 106 & 97 & 82 & 80 & 75 & 64 & 60 & 60 & 59 &  \\
    \midrule 
     \multirow{2}{*}{ostrich} & running & in nest & sitting & riding & red neck & Opening mouth & flapping wings & sleeping & with egg & aside people &  \multirow{2}{*}{548} \\ \cmidrule{2-11}
     & 113 & 91 & 87 & 73 & 34 & 30 & 30 & 30 & 30 & 30 &   \\
    \midrule 
     \multirow{2}{*}{beetle} & longicorn & crawling & on hand & weevil & scarab & ladybird & flying & in hole & on rope & on screen & \multirow{2}{*}{871} \\ \cmidrule{2-11}
     & 137 & 124 & 117 & 111 & 107 & 107 & 78 & 30 & 30 & 30 &    \\
    \midrule 
     \multirow{2}{*}{tent} & mongolia yurt & dome tent & yellow & bell tent & beside bonfire & blue & spire & frame & military & aside people &  \multirow{2}{*}{898} \\ \cmidrule{2-11}
     & 125 & 112 & 110 & 108 & 99 & 96 & 75 & 68 & 62 & 43 &    \\
    \midrule 
     \multirow{2}{*}{kangaroo} & with baby in pouch & jumping & lying & standing & white & grey & tongue out & red & on all fours & fed by human & \multirow{2}{*}{934} \\ \cmidrule{2-11}
     & 174 & 147 & 135 & 131 & 108 & 88 & 60 & 31 & 30 & 30 &    \\
    \midrule 
     \multirow{2}{*}{monkey} & golden monkey & baboon & walking & eating & slow loris & sitting & on rope & on stairs & on shoulder & handstanding & \multirow{2}{*}{881} \\ \cmidrule{2-11}
     & 130 & 127 & 125 & 123 & 121 & 79 & 73 & 43 & 30 & 30 &    \\
    \midrule 
     \multirow{2}{*}{crab} & spotted crab & blue crab & tied up & in hole & belly up & cancer pagurus & in net & on plate & in pot & in hand & \multirow{2}{*}{583} \\ \cmidrule{2-11}
     & 114 & 114 & 94 & 65 & 40 & 36 & 30 & 30 & 30 & 30 &    \\
    \midrule 
     \multirow{2}{*}{lemon} & rotten & half lemon & people eating lemon & on glass & on hand & green & on plate & with fork & in bag & being cutted & \multirow{2}{*}{785} \\ \cmidrule{2-11}
     & 133 & 127 & 116 & 83 & 80 & 79 & 77 & 30 & 30 & 30 &   \\
    \midrule 
     \multirow{2}{*}{pepper} & yellow & orange & green & Chilli & on chopping board & in basket & on plate & half & Spanish paprika & Strip shape & \multirow{2}{*}{767} \\ \cmidrule{2-11}
     & 111 & 108 & 102 & 93 & 80 & 72 & 61 & 58 & 52 & 30 &   \\
    \midrule 
     \multirow{2}{*}{sheep} & lamb & longhorn & on cliff & hairy & sleeping & sheared & on leash & black & aside people & with droopy ears & \multirow{2}{*}{599} \\ \cmidrule{2-11}
     & 117 & 106 & 94 & 54 & 47 & 41 & 39 & 38 & 33 & 30 &    \\
    \midrule 
     \multirow{2}{*}{butterfly} & on hand & swallowtail butterfly & specimen & side view & blue & in cocoon & flying & in glass dome & on mask & on rope & \multirow{2}{*}{872} \\ \cmidrule{2-11}
     & 143 & 130 & 124 & 104 & 98 & 90 & 70 & 53 & 30 & 30 &   \\
    \midrule 
     \multirow{2}{*}{umbrella} & rainbow & hat & long & blue & on hand & in sunlight & folding & on stand & transparent &stowed  &  \multirow{2}{*}{659} \\ \cmidrule{2-11}
     & 121 & 105 & 103 & 57 & 55 & 51 & 47 & 44 & 38 & 38 &    \\
     \bottomrule
\end{longtable}
\end{scriptsize}

\begin{table}[ht]
    \centering
    \caption{The structure of shallow CNNs for MNIST-M}
    \scalebox{0.85}{
    \begin{tabular}{c|c}
        \toprule
        Layer & Details \\
        \midrule
        Input & $3 \times 28 \times 28$ \\
        Conv & Kernel Size 7, Stride 1, Out Channel 32, BN, ReLU \\
        Conv & Kernel Size 5, Stride 2, Out Channel 32, BN, ReLU \\
        Dropout & $p=0.4$\\
        Conv & Kernel Size 3, Stride 1, Out Channel 64, BN, ReLU \\
        Conv & Kernel Size 3, Stride 2, Out Channel 64, BN, ReLU \\
        Dropout & $p=0.4$\\
        FC & Out Channel 16, ReLU\\
        SoftMax & Class\_Num\\
        \bottomrule
    \end{tabular}}
    \tablelabel{tab:shallow}
\end{table}

\begin{figure*}[ht]
    \centering
    \includegraphics[width=1.0\textwidth]{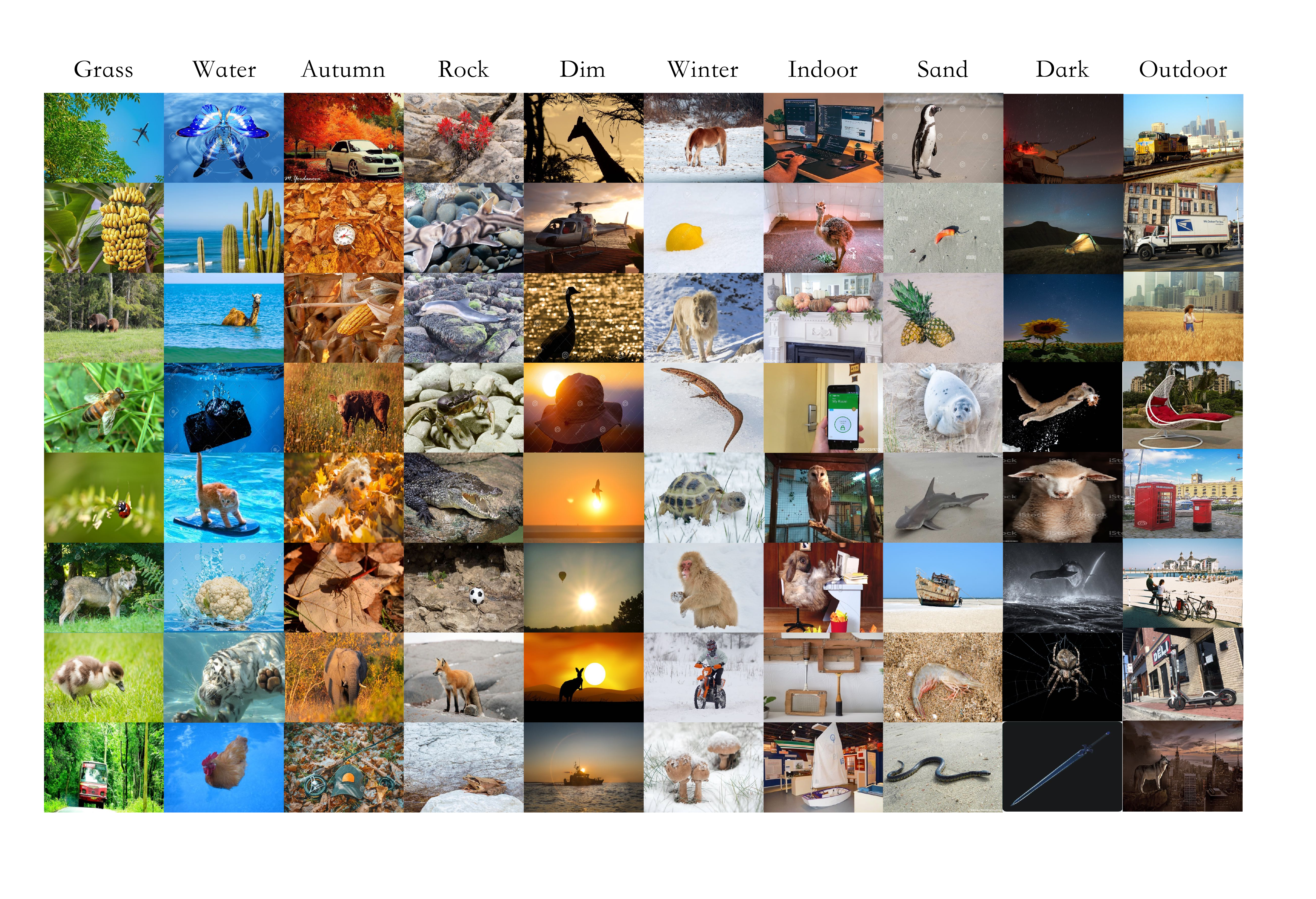}
    \caption{Example Images of common domains in NICO$^{++}$.}
    \figurelabel{fig:common}
\end{figure*}

\begin{figure*}[ht]
    \centering
    \includegraphics[width=0.9\textwidth]{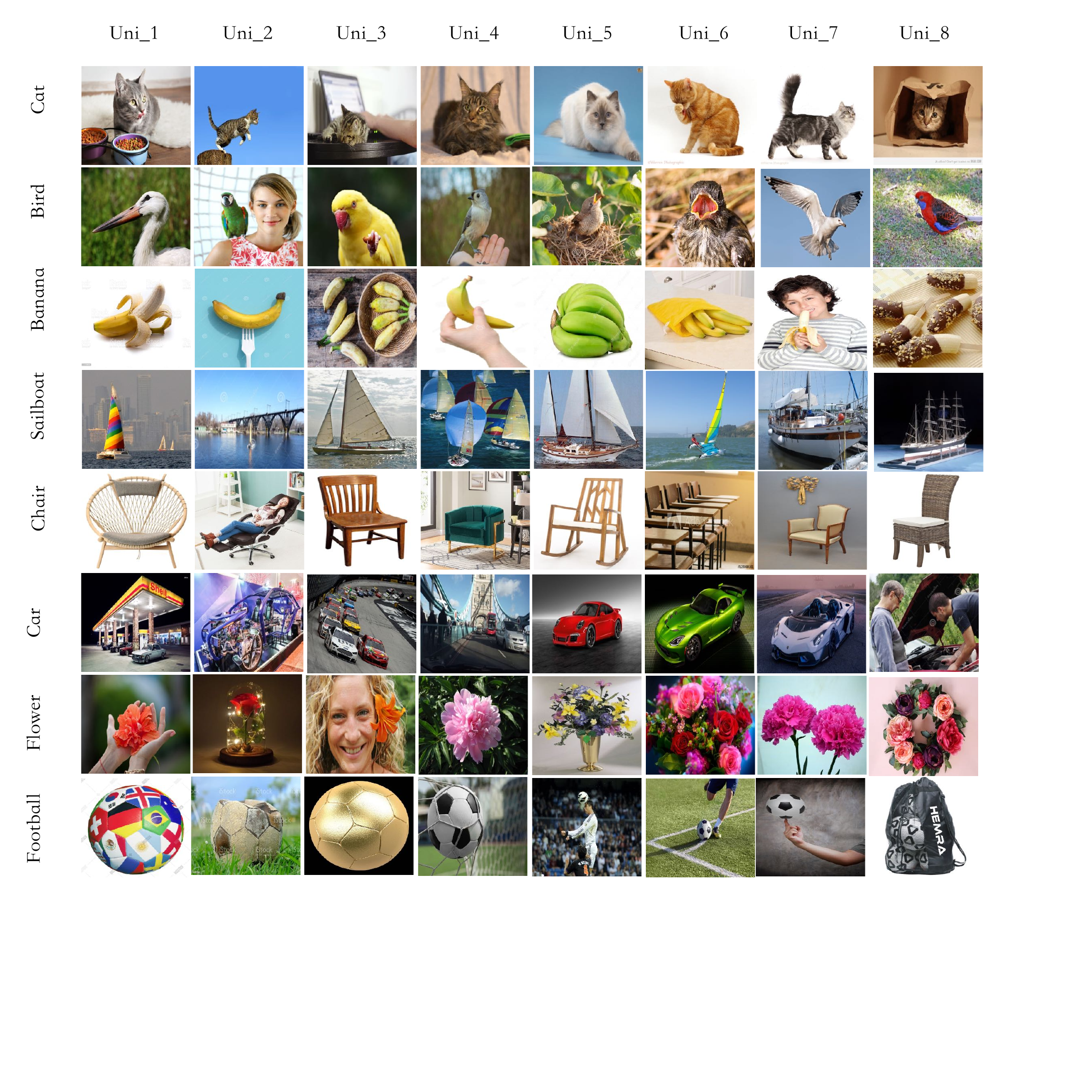}
    \caption{Example Images of unique domains in NICO$^{++}$.}
    \figurelabel{fig:unique}
\end{figure*}

\end{document}